\documentclass[letterpaper]{article} % DO NOT CHANGE THIS
\usepackage{aaai23}  % DO NOT CHANGE THIS
\usepackage{times}  % DO NOT CHANGE THIS
\usepackage{helvet}  % DO NOT CHANGE THIS
\usepackage{courier}  % DO NOT CHANGE THIS
\usepackage[hyphens]{url}  % DO NOT CHANGE THIS
\usepackage{graphicx} % DO NOT CHANGE THIS
\usepackage{natbib}  % DO NOT CHANGE THIS AND DO NOT ADD ANY OPTIONS TO IT
\usepackage{caption} % DO NOT CHANGE THIS AND DO NOT ADD ANY OPTIONS TO IT
\DeclareCaptionStyle{ruled}{labelfont=normalfont,labelsep=colon,strut=off} % DO NOT CHANGE THIS
\usepackage{csquotes}
\usepackage{epigraph} 
\usepackage{enumerate}

\usepackage{balance} % for balancing columns on the final page
\usepackage{xspace}
\usepackage{multirow}
\usepackage{soul} % strike-through
\usepackage{xcolor}
\usepackage{xspace}
\usepackage{amsmath}
\usepackage{subcaption}
\usepackage{cancel}

\usepackage[vlined, ruled, linesnumbered]{algorithm2e} 

\newcommand\algname[1]{\textsf{#1}\xspace}

\newcommand\pbs{\algname{PBS}}
\newcommand\wpbs{\algname{W-PBS}}
\newcommand\tfpbs{\algname{TW-PBS}}
\newcommand\rhcr{\algname{RHCR}}
\newcommand\trhcr{\algname{tRHCR}}

\newcommand{\mapf}{MAPF\xspace}
\newcommand{\tmapf}{tMAPF\xspace}
\newcommand{\mapd}{MAPD\xspace}
\newcommand{\tmapd}{tMAPD\xspace}
\newcommand{\cmapf}{C-MAPF\xspace}

\newcommand{\mediumEnv}{\textsc{Medium}\xspace}
\newcommand{\largeEnv}{\textsc{Large}\xspace}
\newcommand{\largeEnvWide}{\textsc{Large2Wide}\xspace}

\def\A{\mathcal{A}}
\def\D{\mathcal{D}}
\def\O{\mathcal{O}}
\def\G{\mathcal{G}}
\def\V{\mathcal{V}}
\def\E{\mathcal{E}}

\def\T{\mathcal{T}}

\newcommand{\Vstart}{\ensuremath{\V_{\text{start}}}\xspace}
\newcommand{\Vgoal}{\ensuremath{\V_{\text{goal}}}\xspace}

\newcommand{\ignore}[1]{}

\newcommand{\arxiv}[2]{#2}

\title{Terraforming---Environment Manipulation during Disruptions \\ for Multi-Agent Pickup and Delivery}

\author {
    David Vainshtein, %\textsuperscript{\rm 1}
    Yaakov Sherma, %\textsuperscript{\rm 1}
    Kiril Solovey, %\textsuperscript{\rm 1}
    Oren Salzman %\textsuperscript{\rm 1}
    \protect \thanks{
    This research was supported in part by the Israeli Ministry of Science \& Technology grants No. 3-16079 and 3-17385, the United States-Israel Binational Science Foundation (BSF) grants no. 2019703 and 2021643, the Amazon Research Award, and Ravitz Foundation.
    }
}
\affiliations {
    Technion - Israel Institute of Technology, 
    Haifa, Israel
    \\
    dudiwa@cs.technion.ac.il, 
    yaakovsherma@cs.technion.ac.il, 
    kirilsol@technion.ac.il, osalzman@cs.technion.ac.il
}

\begin{document}

\maketitle

\begin{abstract}
In automated warehouses, teams of mobile robots fulfill the packaging process by transferring inventory pods to designated workstations while navigating narrow aisles formed by tightly packed pods. 
This problem is typically modelled as a Multi-Agent Pickup and Delivery (\mapd) problem, which is then solved by repeatedly planning collision-free paths for agents on a fixed graph, as in the Rolling-Horizon Collision Resolution (\rhcr) algorithm. 
However, existing approaches 
make the limiting assumption that agents are only allowed to move pods that correspond to their current task, while considering the other pods as stationary obstacles (even though all pods are movable). This behavior can result in unnecessarily long paths
%, as stationary pods can form narrow corridors and bottlenecks that lead to congestion, 
which could otherwise be avoided by opening additional corridors via \emph{pod manipulation}. 
To this end, we explore the implications of allowing agents the flexibility of  dynamically relocating pods.
We call this new problem Terraforming \mapd (\tmapd) and develop an \rhcr-based approach to tackle it. 
As the extra flexibility of terraforming comes at a significant computational cost, we utilize this capability judiciously by identifying situations where it could make a significant impact on the solution quality. In particular, we invoke terraforming in response to \emph{disruptions} that often occur in automated warehouses, e.g., when an item is dropped from a pod or when agents malfunction. %, which force agents to make even longer detours. 
Empirically, using our approach for \tmapd, where disruptions are modeled via a stochastic process,  we improve throughput by over~$10\%$,  reduce the maximum \emph{service time} (the difference between the drop-off time and the pickup time of a pod) by more than~$50\%$, without drastically increasing the runtime, compared to the \mapd setting. 

\end{abstract}

\begin{figure}[t!]
     \centering
     \begin{subfigure}[b]{0.3\linewidth}
         \centering
         \includegraphics[width=\linewidth]{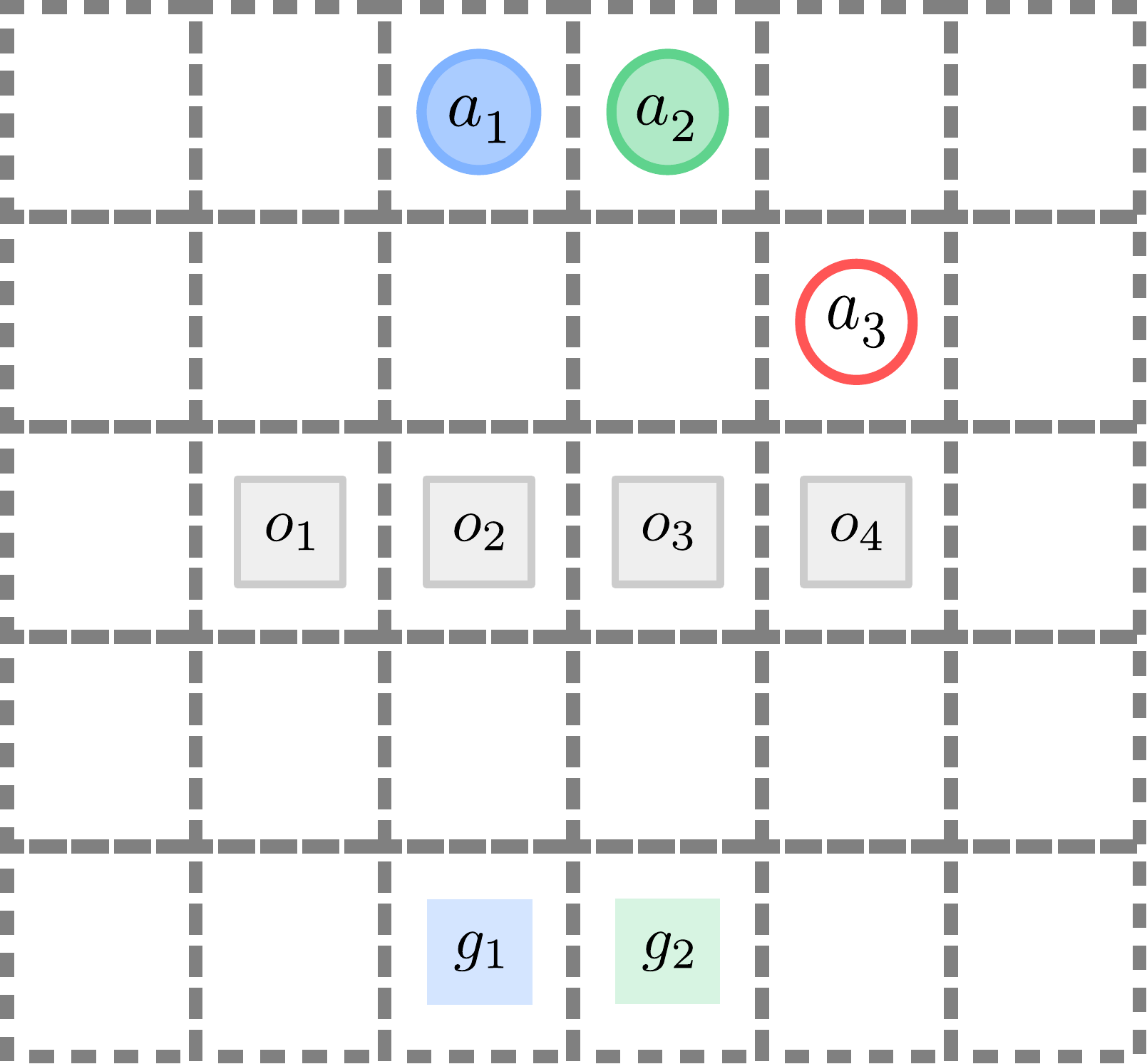}
         \caption{MAPF problem}
         \label{fig:tMAPF1}
     \end{subfigure}
     \hfill
     \begin{subfigure}[b]{0.3\linewidth}
         \centering
         \includegraphics[width=\linewidth]{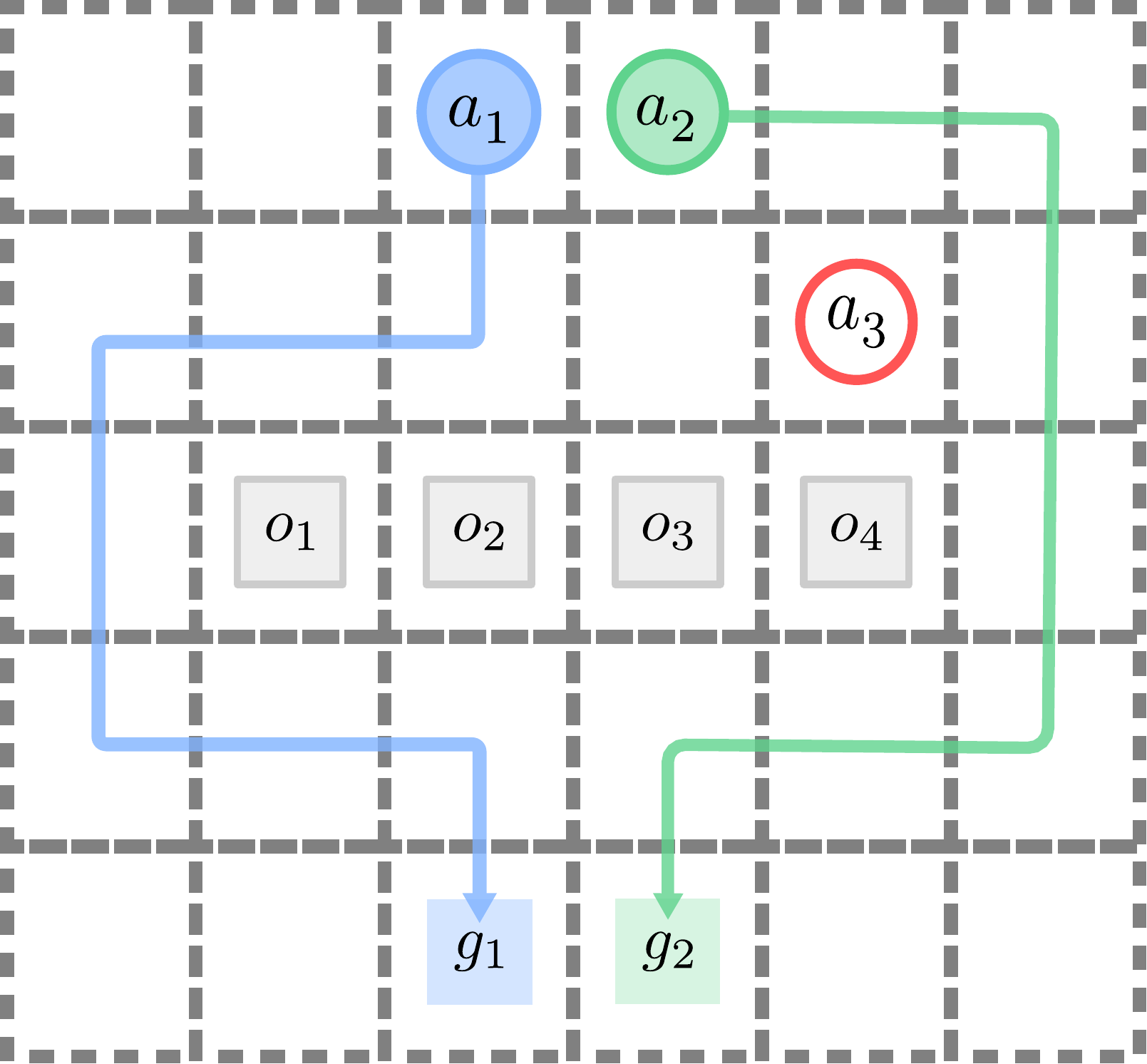}
         \caption{Static obstacles}
         \label{fig:tMAPF2}
     \end{subfigure}
     \hfill
     \begin{subfigure}[b]{0.3\linewidth}
         \centering
         \includegraphics[width=\linewidth]{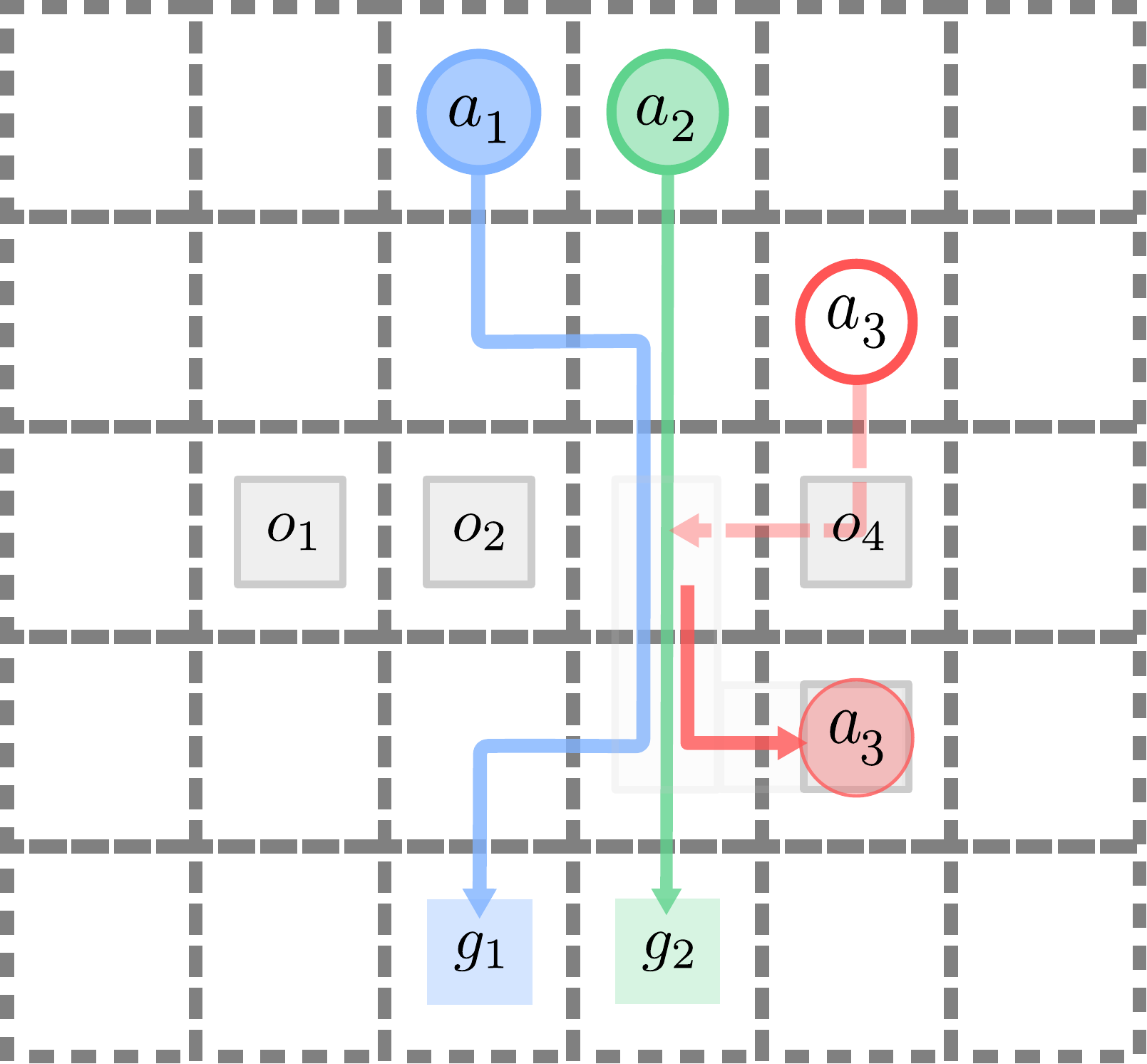}
         \caption{Terraforming}
         \label{fig:tMAPF3}
     \end{subfigure}
     \captionsetup{belowskip=-8pt,aboveskip=8pt}
    \caption{
        Comparing \mapf and \tmapf for a toy problem with agents $\textcolor{blue}{a_1},\textcolor{teal}{a_2},\textcolor{red}{a_3}$ (circles), and a row of obstacles (grey squares). 
        (\subref{fig:tMAPF1})~\mapf problem assigning agents to their goal locations (squares). Here, agent $\textcolor{red}{a_3}$ is a \emph{free} agent which does not carry a pod.
        (\subref{fig:tMAPF2})~\mapf solution where agents must avoid collisions with obstacles and each other. Here, agent~$\textcolor{red}{a_3}$ cannot help the other two agents as all obstacles are static.
        (\subref{fig:tMAPF3})~\tmapf solution where $\textcolor{red}{a_3}$ displaces the movable obstacle $o_3$ to open a passage. 
        %Also see video: \textcolor{blue}{\textsf{https://bit.ly/3u5NELm}}.
        }
        % \vspace{-5mm}
    \label{fig:mapf_intro}
\end{figure}

\section{Introduction}
\label{sec:intro}

Multi-Agent Path Finding (\mapf) is a popular algorithmic framework that captures complex tasks involving mobile agents that need to plan individual routes while avoiding collisions during plan execution~\cite{stern2019multi,SalzmanS20}. 
This abstraction has been successfully applied to a variety of settings (see, e.g.,~\citet{wurman2008cooperative,BelovDBHKW20,LCZCHS0K21,GGSS21,Choudhury.ea.21}). However, in some cases this formulation may not be expressive enough to fully capture the underlying task, which can lead to suboptimal performance. 

This is especially true in the context of automated warehouses where we are given a stream of tasks, and the goal is to maximize the system's throughput.
In this setting, formulated as a  Multi-Agent Pickup and Delivery (\mapd) problem,
inventory pods that hold goods are transported by a large team of mobile agents: 
agents pick up pods, carry them to designated drop-off locations, where goods are manually removed from the pods (to be packaged for customers); each pod is then carried back by a robot to a (possibly different) storage location~\cite{wurman2008cooperative}. {\mapd is typically solved via a sequence of \mapf queries, as in the \rhcr algorithm and its successors~\cite{ma2017lifelong,liu2019task,li2021lifelong,MadarSS22}.
Existing variants of \mapf and \mapd
tend to impose the following limiting and artificial constraint:} pods that are not currently carried to or from a drop-off location are modeled as \emph{static obstacles}, which cannot be moved. Such approaches overlook the fact that pods can be manipulated to clear the way for agents and reduce execution time.

To this end, we explore the implications of allowing agents the extra flexibility of \emph{manipulating the environment by moving obstacles} (i.e., dynamically relocating pod's locations in warehouse applications).
Specifically, we introduce a new \mapd variant (depicted in Fig.~\ref{fig:mapf_intro}) which we term \emph{``Terraforming \mapd''} (\tmapd).\footnote{
The concept of ``Terraforming'' emerged in science fiction at the dawn of the space race, as a means of space exploration, in which a planet's inhospitable environment is altered to facilitate life. For additional details, see, e.g., \url{https://sfdictionary.com/view/125/terraforming}.}
Unfortunately, the extra flexibility of environment manipulation comes at a computational price---solving a \tmapd problem takes significantly longer than a \mapd problem. Roughly speaking, this is because a planner needs to (i)~consider which static obstacles to displace and (ii)~which agent should perform said displacements.
Thus, in this work we suggest to use terraforming only when  unexpected \emph{disruptions} are experienced.

In real-life settings disruptions may occur   when an item is dropped from a pod or when agents experience malfunctions. 
Such events require a safety perimeter to be enforced, cutting off routes and requiring agents to replan in order to detour the disruption area. In the extreme case, agents may become completely trapped by surrounding disruptions, rendering them unable to carry out their tasks.
As we will see, such events are excellent candidates for environment manipulation:
A disruption can be used to guide a planner which obstacles to consider as candidates to be moved allowing the planner to heuristically focus its computation.
More importantly, solutions computed by a planner that can manipulate the environment are often of much higher quality when compared to planners that do not have this extra flexibility making the additional computational effort worthwhile.

We develop an algorithmic approach to tackle the \tmapd problem with disruptions, which casts the question of which obstacles should be moved and by whom to \mapf-like problems. 
Those ideas, which are combined with an \rhcr-based approach, which we call Terraforming \rhcr (\trhcr),  dramatically reduce the size of the huge decision space and allow to efficiently solve  the \tmapd problem.
We evaluate the benefit of \trhcr on warehouse environments that are affected by disruptions, empirically demonstrating an improvement in throughput by over~$10\%$ and a reduction of the maximum \emph{service time} (the difference between the drop-off time and the pickup time of a pod) by more than~$50\%$, without drastically increasing the runtime, as compared with a standard \mapd approach.

\section{Related Work}
Research on \mapf  has produced a wide variety of approaches, ranging from network flow~\cite{YL12}, to
satisfiability~\cite{SFSB16},
answer set programming~\cite{EKOS13}
and tree-search methods~\cite{BSSF14,BFSSTBS15,sharon2015conflict,li2019disjoint}. 
With the focus of this work on the latter group, the method of the most relevance to ours 
%are Conflict-Based Search (\cbs)~\cite{sharon2015conflict} which is used as an algorithmic foundation in many state-of-the-art \mapf solvers (see, e.g.~\citet{GGSS21})  and
is
Priority-Based Search (\pbs)~\cite{ma2019searching} which offers a balance between solution quality and fast computation when solving the \mapd problem~\cite{ma2017lifelong,li2021lifelong}.
In our work we utilize \pbs (detailed in Sec.~\ref{sec:background}) both as our baseline \mapf solver and as the basis for our \tmapd algorithm.

The most closely-related work to our new terraforming problem formulation is by \citet{bellusci2020multi}, in which a configurable environment is optimized alongside path-finding efforts of \mapf. Referred to as the Configurable \mapf (\cmapf) problem, it allows for the reconfiguration of the environment, subject to constraints imposed on the graph itself. An important distinction from our work is that solving the \cmapf problem consists of searching for a \emph{fixed} graph resulting in minimal graph alterations, as well as a set of valid paths dictating where each agent should go.
In contrast, the environment in \tmapd has the capacity to \emph{dynamically change} as agents execute their paths and obstacles are temporarily cleared to make way.

\arxiv{
Other related works~\cite{hauser2013minimum,hauser2014minimum} search for a single-agent path that minimizes the number of obstacles whose removal is necessary to reach the goal without collisions. However, that formulation is not readily extended to a multi-agent setting, nor does it allow agents to remove obstacles.
Recently,~\citet{tang2020computing} introduced the Multi-Robot Clutter Removal (MRCR) problem where a group of agents are tasked with clearing obstacles from a given environment while avoiding collisions with each other and any remaining obstacles. The key differences from our work is that the MRCR formulation allows for obstacles to be removed entirely from the graph, gradually lowering the complexity of the problem as more obstacles are evicted by the agents.}{
}

Also related to our work is a generalization of \mapf called \emph{k-robust} \mapf~\cite{atzmon2018robust} that produces paths guaranteed to be collision-free even when agents are delayed  by up to $k$ timesteps. The concept of robustness is demonstrated as an effective mechanism for avoiding collisions in the context of sudden delays that occur with probability $p$ per each move of each agent \cite{atzmon2020robust}. In our work, although the probability of future per-agent delays $p$ can be modeled, we account for disruptions that block sections of the graph not occupied by an agent and their duration is not known in advance.

Finally, we mention  that the state-of-the-art approach for \mapd is the Rolling-Horizon Collision Resolution (\rhcr) algorithm~\cite{li2021lifelong}. \rhcr iteratively plans a set of partial paths for a group of agents up to a specified time-horizon, decomposing  \mapd  into a series of \mapf queries that are solved iteratively. The choice of \mapf solver is very often \pbs as it offers a balance of solution quality and fast computation time. In our work we extend \rhcr to incorporate terraforming and to account for unexpected disruptions that affect the graph itself.

\section{Problem Formulation}\label{sec:problem}
In this section we start by defining the \mapf problem and the continue to define the Multi-Agent Pickup and Delivery (\mapd) problem and introduce the notion of disruptions in the context of \mapd. Finally, we define the \tmapd problem, which is used in our approach to efficiently handle  disruptions within \mapd settings. 

\subsection{Multi-Agent Path Finding (\mapf)}
We define the \mapf problem as a tuple 
$
\langle
 \G, \A, \Vstart, \Vgoal, \O
\rangle, 
$ where 
the environment $\G=(\V,\E)$ 
is an undirected graph
and $\A=\{a_1,\ldots,a_n\}$ is the set of agents. 
Here,
$\Vstart = \{s_1, \ldots, s_n \}$
and
$\Vgoal = \{g_1, \ldots, g_n \}$ are the agents' start and goal vertices, respectively.
Agents move between vertices along graph edges and are allowed to wait in place. For each agent $a_i$, its actions occur in discrete timesteps of unit cost, resulting in a path $\pi_i$ comprised of a sequence of vertices $\pi_i = \langle s_i, \dots, g_i\rangle$ that is associated with a cost~$|\pi_i|$ of the total number of actions.

A solution to the \mapf problem is a set of collision-free paths $\boldsymbol{\pi} = \{\pi_1, \dots, \pi_n\}$ such that no two agents share the same vertex at the same timestep $\pi_i[t] \neq \pi_j[t]$, nor are they allowed to cross over the same edge in opposing directions $(\pi_i[t], \pi_i[t + 1]) \neq (\pi_j[t + 1], \pi_j[t])$. 
The cost of the solution~$\boldsymbol{\pi}$ is called its \emph{flowtime} and is defined as the sum of individual path costs. Namely,  $|\boldsymbol{\pi}|=\sum_{i}|\pi_i|$ with~$|\pi_i|$ denoting the cost (number of timesteps) of the solution of agent $i$. Note that other cost functions exist, such as  \emph{makespan}~\cite{stern2019multi} that corresponds to the maximum path cost among all agents, i.e.,
$\max\{|\pi_i|\}_{i}$.

In typical \mapf formulations, \emph{static obstacles} that block certain agent positions are implicitly represented by omitting blocked vertices and edges from the graph $\G$. In our setting however, we facilitate interactions between agents and obstacles by explicitly denoting vertices that the agents cannot visit as a set of obstacles $\O = \{o_1, \ldots, o_n \} \subset \V$. 

\subsection{Multi-Agent Pickup and Delivery (\mapd)}
We now describe the standard setting of \mapd where we are tasked with continuously planning for agents as they handle tasks assigned from a task queue $\T$. 
In this work we assume for simplicity that tasks arrive as an online stream, meaning we do not have access to future tasks.

Each task $\tau_i \in \T$ consists of a pair~$\langle p_i, d_i \rangle$  where $p_i\in \O, d_i \in \V$ are its pickup and delivery locations, respectively.
When an agent is assigned a task, it must 
(i)~arrive at the obstacle's pickup location~$p_i$, 
(ii)~carry and deliver it to the delivery location~$d_i$ and 
(iii)~return  it to the pickup location~$p_i$. 
When an agent does not carry an obstacle (i.e., during step~(i) or when it is not assigned with a  task), we say that it is a \emph{free} agent.
When an agent does carry an obstacle (i.e., during steps (ii) and (iii)), we say that it is a \emph{task} agent. 
Let $\tau = \langle p, d\rangle$ be a task assigned to agent $a$.
We define the \emph{pickup time} of $\tau $ as the first time step that $a$ arrives at~$p$.
Similarly, we define the \emph{drop-off time} of $\tau$ as the first time step that $a$ returns to $p$ after delivering the obstacle to $d$ (namely, after the obstacle has been restored to its original location).
The \emph{service time} of $\tau$ in our setting is then defined to be the difference between its drop-off time and its pickup time.

{As we can see, an assigned task $\tau$ has several states: en-route to pickup, delivery and restore. Therefore, it will be convenient to define a \emph{goal mapping} 
$\lambda$ to keep track of each agent's current goal. In other words, $\lambda(a,t,\tau)$ points to the next goal of agent $a$ (being $p$ or~$d$) at timestep~$t$.}

\paragraph{Solution quality.} 
A common measure of solution quality for a \mapd problem is \emph{throughput}~\cite{stern2019multi}, defined as the average number of tasks completed per unit of time, which measures the amortized cost of completing all tasks. 
However, it is often important to consider costs that relate to individual tasks. Thus, we define the \emph{ideal service time} to be the length of the shortest path between a task's pickup and delivery location (and back), while avoiding static obstacles and assuming there are no other agents. The task's \emph{service time ratio}  is then defined as the ratio between the task's actual \emph{service time} and the task's \emph{ideal service time}.
Note that the closer the average service time ratio is to one, the closer the system is to its optimal throughput.

\subsection{Terraforming \mapd with Disruptions}
\label{sec:disruptions}
A \tmapd problem is defined identically to a \mapd problem with the difference that we allow free agents to move obstacles. That is, agents that are not currently carrying obstacles, can pick-up and drop-off obstacles with the purpose of opening passageways and reducing congestion.  
Additionally,  we consider a setting where \emph{disruptions} exist in the environment.
Here, we define a {disruption}~$\D=  \langle v, t_{\rm{start}}, t_{\rm{end}}\rangle$
as the blockage of a vertex $v \in \V$ between timestep $t_{\rm{start}}$ and $t_{\rm{end}}$.
I.e., given such a disruption, no agent can pass through $v$ between $t_{\rm{start}}$ and $t_{\rm{end}}$.
We assume that disruptions are unpredictable: 
at every timestep $t$ the planner only has access to the currently active disruptions through a function \textsc{Observe}. The function specifies all vertices $\V_{\D}$ that are blocked at $t$ (i.e.,  vertices of disruptions where $t \in [t_{\rm{start}}, t_{\rm{end}}]$). 
This implies that the \mapd planner does not have access to (i)~future disruptions or (ii)~the time $t_{\rm{end}}$ for which an  active disruption will end. 
In our setting, disruptions occur only along paths traversed by agents thus modelling realistic disruptions in warehouses such as items dropped from a pod or when agents malfunction.\footnote{Our disruption model shares similarities to existing models in other domains such as railway planning~\cite{mohanty2020flatland}.}

\section{Algorithmic Background}
\label{sec:background}
In preparation to our approach for \tmapd, we first describe the \pbs algorithm as a solver for the (classical) \mapf problem. We then describe how \pbs can be used to solve the (standard) \mapd problem, as is often done in practice.

\subsection{Priority-Based Search}
We  provide an overview of \pbs together with an adaptation termed windowed \pbs (\wpbs{}) for \mapf and refer the reader to~\citet{ma2019searching} for further details.
Pseudocode of \wpbs{} is detailed in Alg.~\ref{alg:tfpbs}.\footnote{
Blue text in Alg.~\ref{alg:tfpbs}  will be used to explain the adaptation of \wpbs{} to terraforming in Sec.~\ref{sec:algs} and should be ignored for now.
}
At its core
%, \pbs resembles \cbs, as it 
\pbs takes a hierarchical approach using a  high- and low-level search in which \pbs maintains priorities between agents at the high-level and searches for agent paths that abide to these priorities in the low-level. 
%However, unlike \cbs which maintains space-time constraints between the agents in the high-level search, \pbs maintains priorities between agents.

More specifically, in the high-level search, \pbs explores a \emph{priority tree} (PT), where a given node~$N$ of the PT encodes a (partial) priority set $P_N=\{a_h \prec a_i,~a_j \prec a_l$,~\ldots \}. A priority $a_{i}\prec a_{j}$ means that agent $a_{i}$ has precedence over agent~$a_{j}$ whenever a low-level search is invoked (see below).
In this case, we say that~$a_i$ has a higher priority than~$a_j$.
In addition to the ordering, each PT node maintains single-agent paths that represent the current \mapf solution (possibly containing collisions.
The \pbs algorithm (Alg.~\ref{alg:tfpbs}) starts the high-level search with the tree root whose priority set is empty, and assigns to each agent its shortest path (Lines~$2$-$6$). Whenever \pbs expands a node $N$ (Line~$8$), it invokes a low-level search to compute a new set of paths which abide by the priority set $P_N$. If a collision between agents, e.g.,~$a_i$ and~$a_j$, is encountered in the new paths, \pbs generates two child PT nodes $N_1, N_2$ with the updated priority sets $P_{N_1}=P_N\cup \{a_i\prec a_j\}, P_{N_2}=P_N\cup \{a_j\prec a_i\}$, respectively (Line~$12$). The high-level search chooses to expand at each iteration a PT node in a depth-first search manner. 
The high-level search terminates when a valid solution is found at some node $N$ (Line~$10$), or when no more nodes for expansion remain, in which case, \pbs declares failure. 

The low-level search of \pbs proceeds in the following manner. For a given PT node $N$, \pbs performs a topological sort of the agents according to $P_N$ from high priority to low, and plans individual-agent paths based on the ordering. 
For a given topological ordering $(a'_{1},\ldots, a'_{k'})\subset \A$, for some $1\leq k'\leq k$, the low-level iterates over the~$k'$ agents in the topological ordering, and updates their paths such that they do not collide with any higher-priority agents. Note that agents that do not appear on this list maintain their original plans. It then checks all agents for any remaining collisions.

As we will see shortly, to speed up planning times, it is often convenient to consider collisions only for the first $\omega$ timesteps.  
As previously mentioned, this variant is called \emph{windowed}-\pbs{} or \wpbs{} for short.

\begin{algorithm}[t]
\SetAlgoVlined
\DontPrintSemicolon
\SetNoFillComment
\SetKwInput{KwData}{Input}
\SetKwInput{KwResult}{Returns}
\KwData{Graph $\G$, agents $\A$, \textcolor{blue}{movable obstacles $\tilde{\O}$},
\newline
{goal mapping} $\lambda$,
% \newline
planning window $\omega$}
\KwResult{A collision-free plan $\boldsymbol{\pi}$}
  \textcolor{blue}{$\A \gets \A \cup \tilde{\O}~~~/\!\!/\texttt{\:treat}~\tilde{\O}~\texttt{\:as\:demi-agents}$\;}
 $R.priorities \gets \emptyset 
 \quad~/\!\!/\texttt{\:root\:state}$\;
 $R.paths \gets \mathrm{findPaths}(\G,~\A,~\lambda,~R.priorities, \omega)$\;
 $R.cost \:\:~\gets \mathrm{get}\textcolor{blue}{\mathrm{Terra}}\mathrm{Flowtime}(R.paths)$ \;
 $R.collisions \gets \mathrm{detectCollisions}(R.paths)$\;
 $\mathrm{insert}(R, \textsc{Open})$\;
 \vspace{4pt}
 \While{$\textsc{Open}$ not empty}{
  $N \gets \mathrm{pop}(\textsc{Open})~~/\!\!/\texttt{\:searched\:via\:DFS}$\;
  $\langle a_i, a_j, l, t \rangle \gets \mathrm{getCollisions}(N)$\;
  \vspace{4pt}
  \If{$N.collisions$ is empty}{
    \Return $N.paths$
  }
  \vspace{4pt}
  \For{$p \in \langle a_i \prec a_j \rangle, \langle a_j \prec a_i \rangle$}{
        $P \!\:~\gets N.priorities \cup \{p\}$\;
        $N^\prime \gets \mathrm{clone}(N)$\;
        $N^\prime.paths \gets \mathrm{findPaths}(\G,~\A,~\lambda,~P,~\omega)$\;
        $N^\prime.cost \gets \mathrm{get}\textcolor{blue}{\mathrm{Terra}}\mathrm{Flowtime}(N^\prime.paths)$\;
        $N^\prime.collisions \gets \mathrm{detectCollisions}(N^\prime.paths)$\;
        $N^\prime.priorities \gets P$ \;
        $\mathrm{insert}(N^\prime, \textsc{Open})$\;
      }
  }
 \caption{\textcolor{blue}{T}W-PBS}
 \label{alg:tfpbs}
\end{algorithm}

\subsection{Solving \mapd using \rhcr
}
\label{sec:mapdapp}
The common approach to solve \mapd problems~\cite{li2021lifelong, okumura2021iterative} is by iteratively
(i)~assigning tasks to free agents,
(ii)~deriving target locations from these tasks and setting starting locations to be the agents' current locations
and
(iii)~running a \mapf solver to compute collision-free paths for all the agents.
The first step, task assignment, can be solved in a variety of methods (see, e.g.,~\cite{ma2017lifelong,ma2019lifelong}) but in this work we limit ourselves to greedy task assignment. Specifically, we compute the graph distance (i.e., the number of edges in a shortest path) of each agent to each goal and assign tasks greedily according to these distances.
The last step, solving a MAPF problem, is typically done by computing collision-free paths up to a certain time-horizon hence running a windowed \mapf solver such as \wpbs. These aspects form the \rhcr  algorithm~\cite{li2021lifelong} which demonstrates state-of-the-art performance by re-planning agent paths in regular periods of $h$ simulation timesteps and resolving inter-agent collisions occurring within a time window $w$, such that $w \geq h$. The use of bounded-horizon \pbs (i.e., \wpbs) yields high throughput at a reduced computation effort, albeit without completeness or optimality guarantees.

{A detail that is crucial to iteratively applying \wpbs as tasks arrive in an online manner, is the careful handling of priority ordering. Namely, }whenever an agent is assigned a new task, or completes a sub-task, its path is replanned with its priority ordering wiped, to ensure it will not give automatic precedence over other agents based on past interactions. After each time step, the planning horizon is extended and paths are replanned until all tasks are complete.

\section{Algorithmic Framework}
\label{sec:algs}
In this section we present our algorithmic framework for incorporating terraforming into \mapd. 
Our approach follows the  \rhcr approach (Sec.~\ref{sec:mapdapp}) to solve \mapd, i.e., we assign new tasks to agents and then decompose the problem into a sequence of \mapf instances to use \wpbs to solve these individual queries. 
Subsequently,  we observe the environment to detect if there are disruptions. 
When disruptions are observed, we start by planning new paths for agents (without terraforming). 
We then consider if it is worthwhile to invoke terraforming to manipulate the environment. 
To do so, we initiate a simplified terraforming \mapf (\tmapf) problem (defined below) where the set of movable obstacles is in the local neighborhood of the disruption, and the movable obstacles are \emph{self-propelled}, i.e., can move without the intervention of agents.
This simplification
allows us to defer the question of who manipulates an obstacle
to a later stage, and momentarily only reason
about whether obstacles should be moved at all.
The solution to this \tmapf problem does not fully account for the cost of moving the obstacles and is used to evaluate which obstacles (if any) should be displaced. Subsequently, new tasks are defined and assigned to agents corresponding to obstacles that should be displaced as specified by the \tmapf solution. 

For simplicity, we assume that terraforming tasks are structured similarly to standard tasks (i.e., there is a delivery location to which the obstacle will temporarily be moved to). Thus, in addition to the pickup location of each task, which corresponds to the location of an obstacle, a delivery location must be determined. To this end, we assume there is a defined subset of vertices within the graph $\hat{\V}\subset \V$, exclusively reserved for this purpose. When a new terraforming task is created, the nearest vertex from this subset is selected as the task’s delivery location.

The rest of the section formalizes and details our approach.
Specifically, we start with detailing the high-level approach for \tmapd, which we call \trhcr, and then continue to describe the individual components such as an adaptation of \wpbs{} to terraforming.

\begin{algorithm}[t]
\SetAlgoVlined
\DontPrintSemicolon
\SetNoFillComment
\SetKwInput{KwData}{Input}
\SetKwInput{KwResult}{Returns}
\KwData{Graph $\G=(\V,\E)$, 
{affected} agents $\tilde{\A}$, \qquad self-propelled obstacles $\Tilde{\O}$, {goal mapping} $\lambda$, \newline
planning window~$\omega$,
\newline
terraforming reserved locations $\hat{\V}$}
\KwResult{A set of Terraforming tasks $\Tilde{\T}$}
$\pi_{\textsc{movable}} \gets \tfpbs(\G,
\Tilde{\A},\Tilde{\O}, \lambda, \omega)$ \;
$\Tilde{\O}_{\textsc{moved}} \gets \mathrm{getObstaclesToMove}(\pi_{\textsc{movable}})$\;
$\Tilde{\T} \gets {\mathrm{convertToTasks}(\Tilde{\O}_{\textsc{moved}}, \hat{\V})}$\;
return $\Tilde{\T}$\;

  \caption{{getTerraformingTasks}}
 \label{alg:terra-tasks}
\end{algorithm}

\subsection{Terraforming \rhcr}
\label{subsec:approach}
We are ready to detail our algorithmic framework \trhcr for \tmapd (Alg.~\ref{alg:tf-mapd}).
Recall that at every timestep and while tasks remain (Line~$2$), we assign tasks to agents (Line~$3$) and replan paths for agents that obtain new tasks or agents whose path requires re-planning to ensure a sufficient planning horizon (Line~$4$).
We then observe for disruptions (Line~$5$), which is where our approach deviates from \rhcr.\footnote{Note that if path planning took place after observing new disruptions, all agents would avoid these disruptions. By planning paths before checking for disruptions, agents are unaware of new obstructions and assume their desired path will be available allowing us to identify which agents are affected by new disruptions.}

\begin{algorithm}[t]
\SetAlgoVlined
\DontPrintSemicolon
\SetNoFillComment
\SetKwInput{KwData}{Input}
\SetKwInput{KwResult}{Returns}
\KwData{Graph $\G=(\V,\E)$, agents $\A$, obstacles $\O$, 
tasks  stream $\T$, terraforming radius $r$, planning window~$\omega$, \\
\qquad \quad terraforming reserved locations $\hat{\V}$}
\KwResult{A collision-free \mapd plan $\boldsymbol{\pi_{\textsc{mapd}}}$}
 $t \gets 0$\;
 \While{$\T$ is not empty \textbf{or} agents have tasks}{
    $\lambda \gets \mathrm{assignTasks}(\A,\T)$\;    

    $\boldsymbol{\pi} \gets \wpbs(\G,\A,\lambda, \omega)$\;
    \vspace{4pt}
    $\V_{\D} \gets \textsc{Observe}(\G, t)$\;
    \If{$\V_{\D}$ is not empty}{
        $\Tilde{\A} \gets \mathrm{getAffectedAgents}(\boldsymbol{\pi},~ \V_{\D})$\;
        $\boldsymbol{\pi} \gets \wpbs(\G \setminus \V_{\D},\Tilde{\A},\lambda, \omega)$\;

        $\Tilde{\O} \gets \mathrm{candidateObstacles}(\O,~ \V_{\D}, r)$\;
        ${\Tilde{\T} \gets {\mathrm{getTerraformingTasks}(\G, \Tilde{\A},\Tilde{\O},\lambda,\omega, \hat{\V}})}$\;
        $\tilde{\lambda}\gets \mathrm{assignTasks}(\A,
        {\tilde{\T}\cdot \T})$\;
        $\boldsymbol{\pi_{\textsc{terra}}} \gets \wpbs(\G,\A,\tilde{\lambda},\omega)$\;
        \vspace{4pt}
        \If{$|\boldsymbol{\pi_{\textsc{terra}}}| < \boldsymbol{|\pi}|$}{
            $\boldsymbol{\pi} \gets \boldsymbol{\pi_{\textsc{terra}}}$\;
            $\T \gets {\Tilde{\T}\cdot\T}$\;
        }
      }
  $\boldsymbol{\pi_{\textsc{mapd}}}$.append actions from $\boldsymbol{\pi}$ at timestep $t$ \;
  $t\gets t+1$\;
  }
  \caption{{\trhcr}}
 \label{alg:tf-mapd}
\end{algorithm}

\begin{figure*}[bt]
     \centering
     \begin{subfigure}[b]{0.18\linewidth}
         \centering
         \includegraphics[width=0.9\linewidth]{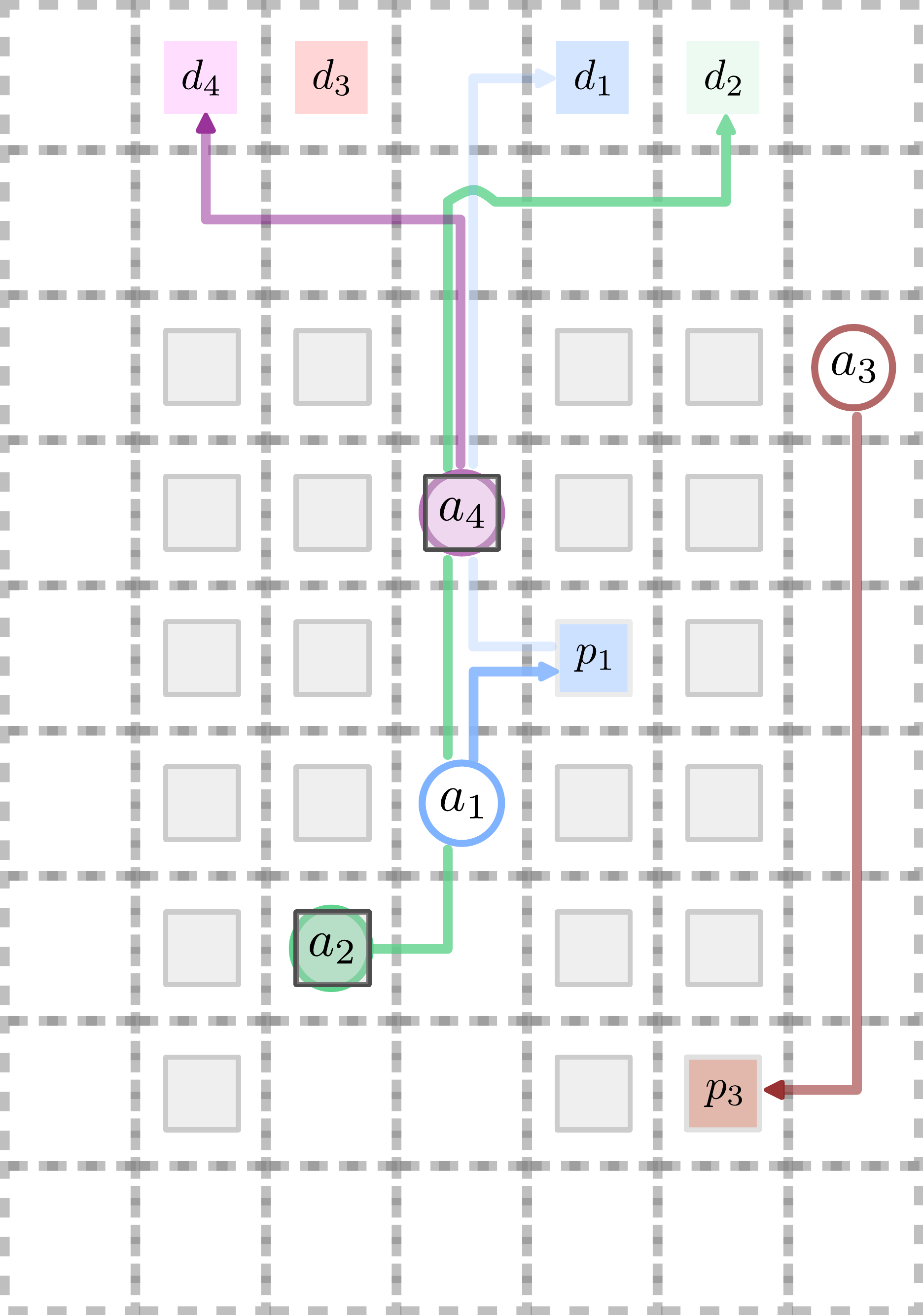}
         \caption{$t$}
         \label{fig:tMAPD1}
     \end{subfigure}
     \hfill
     \begin{subfigure}[b]{0.18\linewidth}
         \centering
         \includegraphics[width=0.9\linewidth]{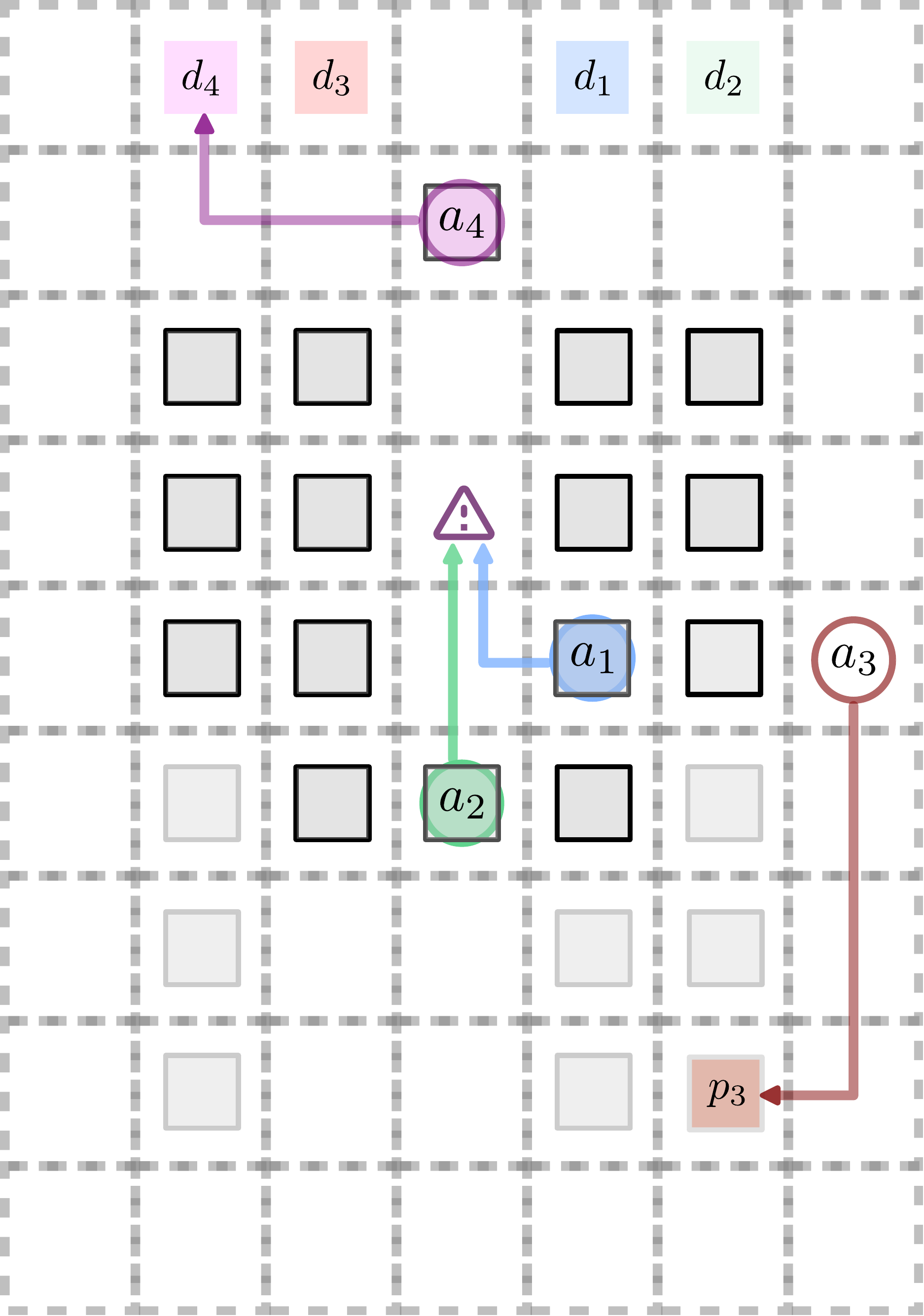}
         \caption{$t+2$}
         \label{fig:tMAPD2}
     \end{subfigure}
     \hfill
     \begin{subfigure}[b]{0.18\linewidth}
         \centering
         \includegraphics[width=0.9\linewidth]{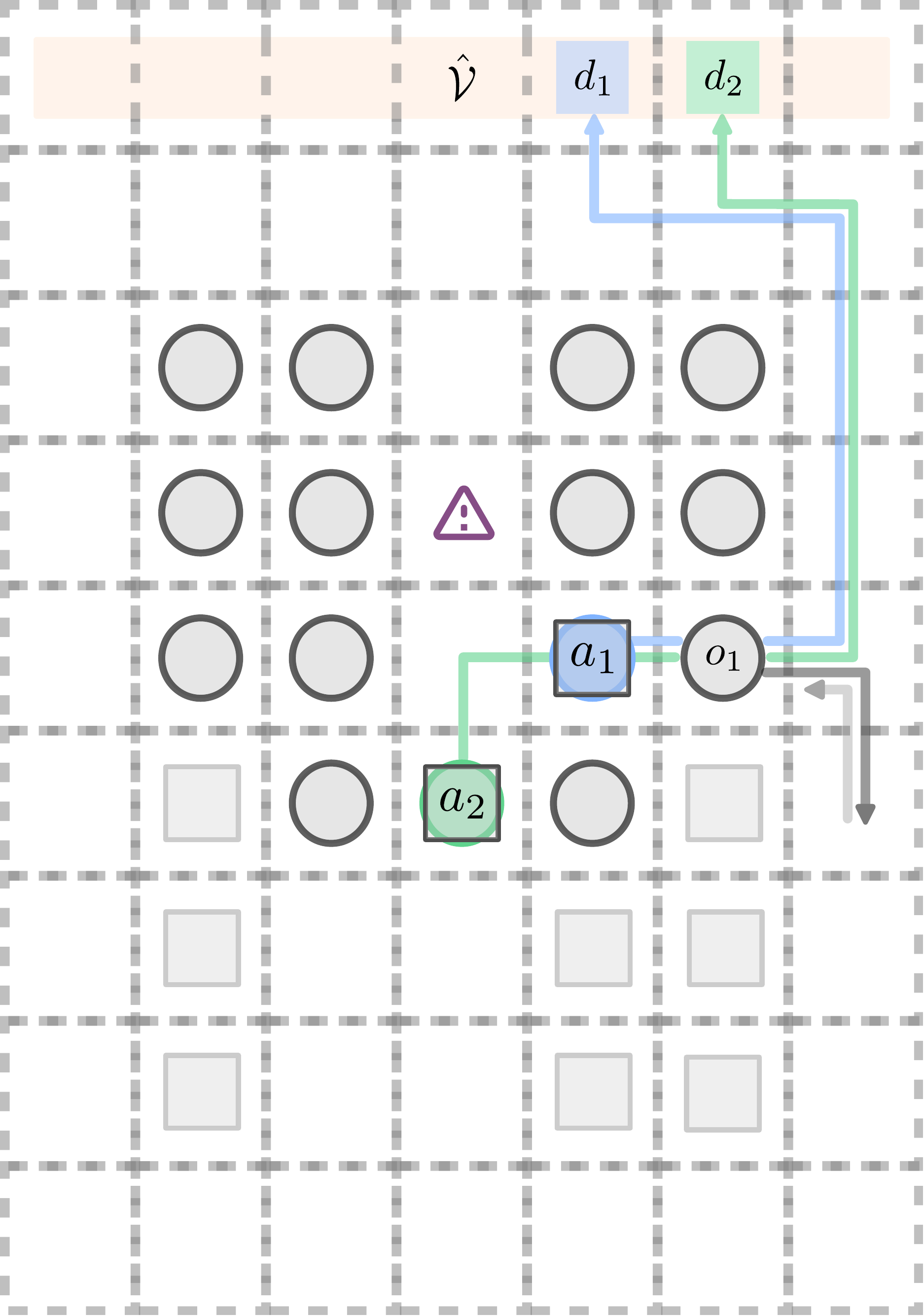}
         \caption{tMAPF instance}
         \label{fig:tMAPD3}
     \end{subfigure}
     \hfill
     \begin{subfigure}[b]{0.18\linewidth}
         \centering
         \includegraphics[width=0.9\linewidth]{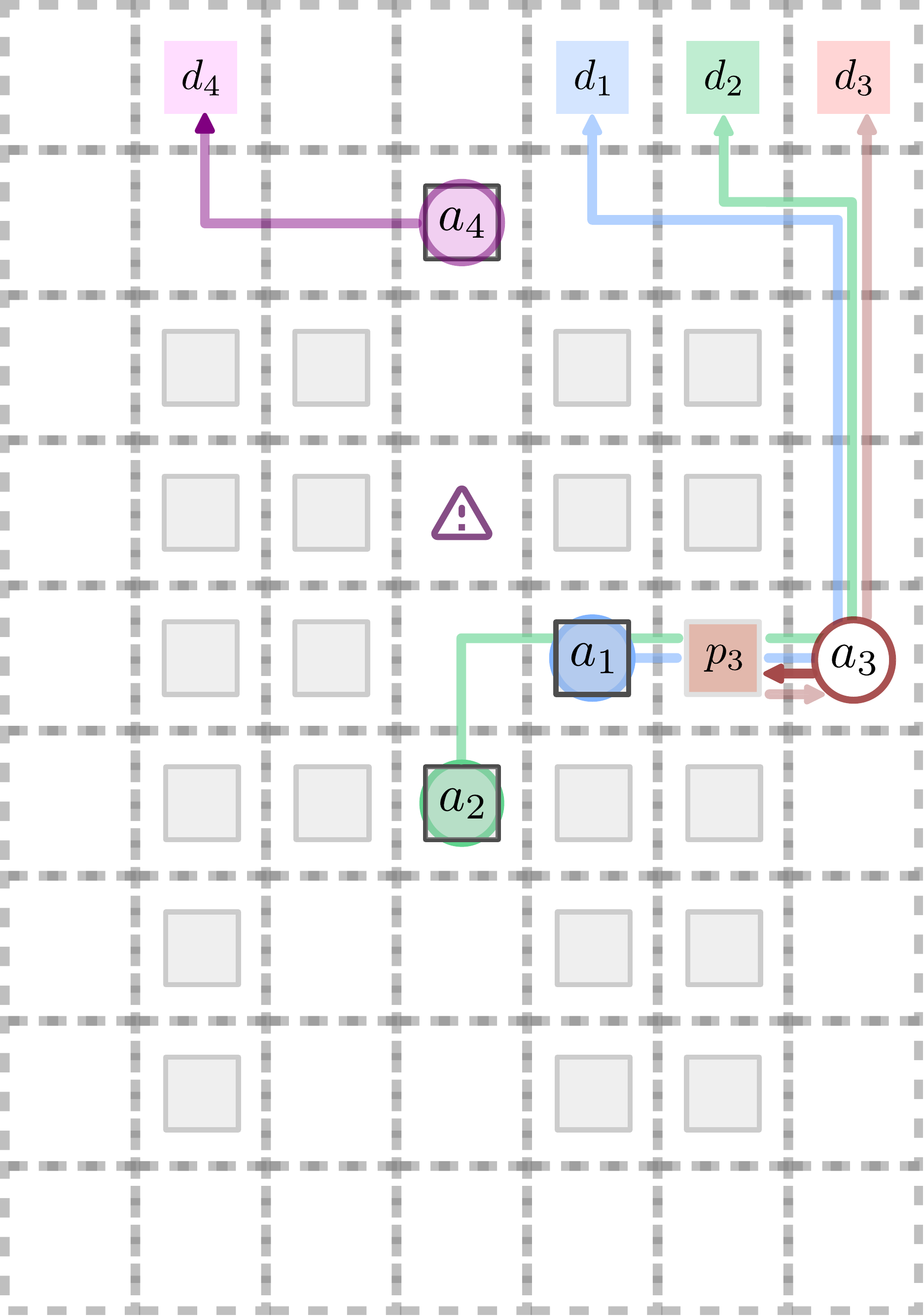}
         \caption{$\boldsymbol{\pi_{\textsc{terra}}}$}
         \label{fig:tMAPD4}
     \end{subfigure}
     \hfill
     \begin{subfigure}[b]{0.18\linewidth}
         \centering
         \includegraphics[width=0.9\linewidth]{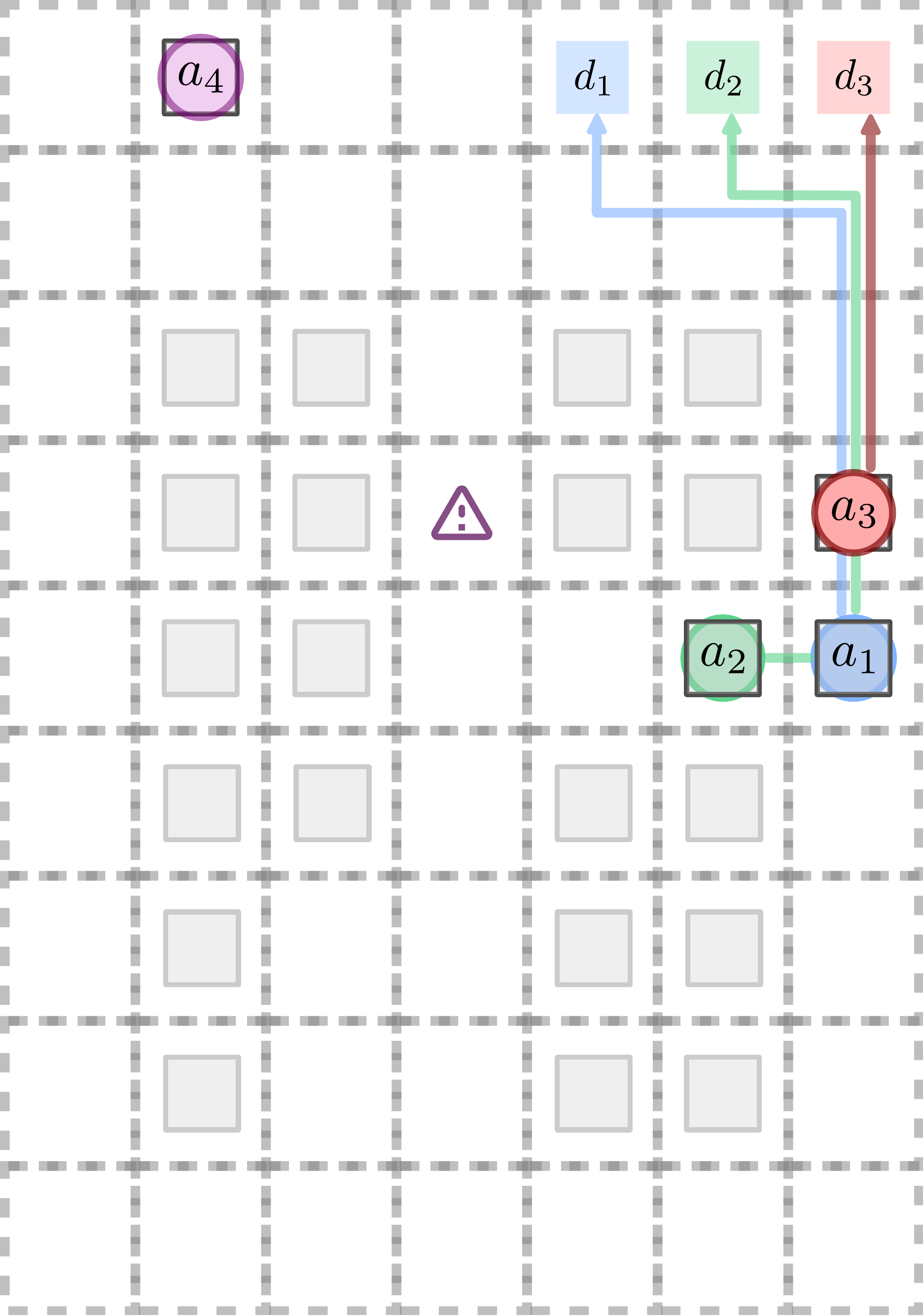}
         \caption{$t + 5$}
         \label{fig:tMAPD5}
     \end{subfigure}
     \captionsetup{belowskip=-8pt,aboveskip=8pt}
    \caption{
        \trhcr visualization (see description in Sec.~\ref{subsec:alg-details}).
        }
        % \vspace{-5mm}
    \label{fig:MAPD_example}
\end{figure*}

As mentioned, disruption detection is done using the  \textsc{Observe} function, which returns a set~$\V_{\D} \subset \V$ of all {newly detected} disruption locations at the current timestep $t$ (recall that we do not have access to the termination  time of the disruptions). 
When disruptions occur (Lines~$6$-$15$), 
we start by computing the set of agents~$\tilde{\A}$ that are \emph{affected} by the disruptions using the routine \rm{getAffectedAgents} (see
\arxiv{Alg.~\ref{alg:get-affected} and}{}%
details below). 
Informally, an agent $a_i$ is considered to be affected if its path gets blocked due to a disruption. 
Additionally, we include any agent $a_j$ that gives $a_i$ precedence (i.e., $a_i \prec a_j$) as $a_j$ may have altered its desired path due to a restriction imposed by $a_i$'s path.
We then call \wpbs (Line~$8$) to replan paths for the affected agents while preserving the paths of those unaffected.
%(unless they become involved in new collisions). 
Note that Terraforming does not take place at this stage.

To consider Terraforming, we start (Line~$9$) by computing a set of candidate obstacles~$\tilde{\O}$ that  {are evaluated} for displacement. 
The set $\tilde{\O}$ is defined to be all obstacles within a graph distance of $r\geq 1$ from a {disruption}, where~$r$ is called the \emph{terraforming radius}.\footnote{In the case of a grid, the graph distance (the number of edges in a shortest path connecting two vertices) is the Manhattan distance.}
Namely,
$$
\tilde{\O} := 
\{
o \in \O ~ \vert ~ \exists v \in \V_{\D}~\text{ s.t. }~\| o,v\| \leq r
\}.
$$
In the next step, we identify which obstacles from $\tilde{\O}$ could be potentially displaced 
in order to clear the way for the affected agents~$\tilde{\A}$ and where those should be displaced to. This is done by calling the $\mathrm{getTerraforimingTasks}$ subroutine in Line~$10$ (detailed in Sec.~\ref{subsec:alg-mapf}), which returns a new set of tasks $\tilde{\T}$ that specify which obstacles from $\tilde{\O}$ should be moved, and where to move it. 

% \modified{Note that $\mathrm{getTerraforimingTasks}$ takes as input also the set of reserved locations which are used for terraforming (exaplanied in Sec.~\ref{subsec:alg-mapf}).}
%
In the next step (Line~11),
we compute a new assignment~$\tilde{\lambda}$ for the agents~$\A$.
Here we assign both the new tasks~$\tilde{\T}$ which may require agents to move obstacles in $\tilde{\O}$ and the original tasks~$\T$ (we use $\tilde{\T}\cdot \T$ to denote the concatenation of the two task sets).
Note that the new tasks~$\tilde{\T}$ are assigned before the existing tasks $\T$ and that only the first~$\vert \A \vert$ tasks are assigned.
After the new assignment was defined, a new \mapf problem is defined and solved (Line~12) to obtain a solution~$\boldsymbol{\pi_{\textsc{terra}}}$. Finally, we choose $\boldsymbol{\pi}$ to be the lowest-cost solution among the options of not using and using terraforming (Lines~13-14).

\subsection{Details for getTerraformingTasks}
\label{subsec:alg-mapf} 
We provide details on the $\mathrm{getTerraformingTasks}$ subroutine. First, we describe the \tmapf instance it solves within, our solution approach for \tmapf, which we call \tfpbs, and the steps in Alg.~\ref{alg:terra-tasks}.

\paragraph{\tmapf.}
%\tmapf allows the planner to alter the environment in order to open passageways and create shortcuts.
We define the \tmapf problem as a generalization to the \mapf problem, 
% where in addition to the graph~$\G$ and agents~$\A$, 
where in addition to the graph~$\G$, agents~$\A$ and goal mapping~$\lambda$,
the input  includes a subset of obstacles~$\tilde{\O}\subset\O$ as movable obstacles. 
An obstacle $o\in \tilde{O}$ needs to be restored to its original position to avoid a permanent alteration of the environment.
Recall that we make the simplifying assumption that the movable obstacles~$\tilde{\O}\subset\O$ are \emph{self-propelled}, i.e., they can move on their own if necessary.% instructed to do so. 
Hence we will refer to them as ``demi-agents''. 

Intuitively, demi-agents are considered as already waiting at their goal location and regular agents may collide with them. This will cause either the colliding agent to recompute its path or cause the demi-agent to  move in order to allow the agent to pass. 
The difference between costs with and without terraforming  lies in how we compute the cost of a demi-agent's path.
In contrast to a (regular) agent's path~$\pi$ whose cost~$\vert\pi\vert$ is the number of timesteps taken to execute~$\pi$,
for a demi-agent's path~$\tilde{\pi}$, we define its cost  $\vert\tilde{\pi}\vert$ as the number of movements taken by the demi-agent (i.e., wait actions do not incur a cost, so that idle obstacles do not incur a penalty cost). Without this modification, performing terraforming would not be worthwhile as all movable obstacles would incur cost, regardless of whether they moved or not. 
To this end, with a slight abuse of notation, we denote the cost of a solution as 
$\vert \boldsymbol{\pi} \vert =\sum_{i\in\A}|\pi_i| + \sum_{j\in\tilde{O}}|\tilde{\pi}_j|$
and refer to it as the \emph{terra-flowtime}.

\paragraph{Solving \tmapf with \tfpbs.}

Alg.~\ref{alg:tfpbs} outlines the pseudocode of \tfpbs with the differences from \wpbs highlighted in blue.
Specifically, the set of movable obstacles $\tilde{\O}$ are treated as ``demi-agents'' and added to the set of agents $\A$ to create the set of agents the algorithm considers (Line~$1$).
When the cost of a node is computed (Line~$4$ and~$16$), instead of computing the (standard)  flowtime via the function \rm{getFlowtime}, we compute the terraforming \emph{flowtime} via the function \rm{getTerraFlowtime} that does not account for movable obstacles' wait actions. 

\paragraph{Wrapping up.}
We finalize the details of Alg.~\ref{alg:terra-tasks}. 
We start (Line~1) by solving the \tmapf problem that allows obstacles to move upon collision with agents and let $\pi_{\textsc{movable}}$ be the solution to this problem. 
% \modified{The input to the \tmapf query includes the graph $\G$, the set of affected }
%
We then extract the set of obstacles~$\Tilde{\O}_{\textsc{moved}}$ that are required to move as part of~$\pi_{\textsc{movable}}$ (Line~2)
and use them to define a new set of tasks~$\Tilde{\T}$ (Line~3), defined as follows.
For each obstacle that needs to be moved~$o\in \Tilde{\O}_{\textsc{moved}}$, we generate a new task $\langle o, \ell \rangle$, 
where~$o\in\O$ (the obstacle's location) is the task's pickup location and $\ell\in \hat{\V}$ is the delivery location.
$\ell$ is chosen to be the point nearest to the pickup location~$o$ in the reserved location set.
Upon being assigned the task, an agent will move the obstacle to perform terraforming, before returning it to its original location. 
The newly generated set of tasks~$\Tilde{\T}$ is then returned as output (Line~4).  
Note that the computed path of a self-propelled obstacle in the \tmapf problem is not used, but rather it only serves as a hint on whether it pays off to move the obstacle in the first place. 

% Note that the actual paths of the agents in the \tmapf problem, are not used. We only use its solution to decide which obstacles should be moved.
\arxiv{
\begin{algorithm}[t]
\SetAlgoVlined
\DontPrintSemicolon
\SetNoFillComment
\SetKwInput{KwData}{Input}
\SetKwInput{KwResult}{Returns}
\KwData{Agent paths $\boldsymbol{\pi}$, set of active disruption locations $\V_{\D} \subset \V$}
\KwResult{Set of affected agents $\tilde{\A}$}
 $\tilde{\A} \gets \{a_i \vert \text{ path } \pi_i \in \boldsymbol{\pi} \text{ of } a_i \text{ intersects vertex in } \V_{\D}\}$\;
 $\mathrm{push}(S, \tilde{\A})$~~$/\!\!/\texttt{\:push\:agents\:into\:stack\:}S$\;
 \While{$S$ is not empty}{
    $a \gets \mathrm{pop}(S)$\;
    % (\modified{\texttt{/\!\!/\:$a$ has higher priority}})
     \For{$a' \textbf{ s.t. } a \prec a'$}{
        \If{$a' \notin \tilde{\A}$}{
            $\mathrm{push}(S, a')$\;
            $\tilde{\A} \gets \tilde{\A} \cup \{a'\}$\;        }
      }
  }
  return $\tilde{\A}$\;
 \caption{getAffectedAgents}
 \label{alg:get-affected}
\end{algorithm}
}{}

\begin{figure*}[t]
     \centering
     \begin{subfigure}[b]{0.4\columnwidth}
         \centering
         \includegraphics[width=\columnwidth]{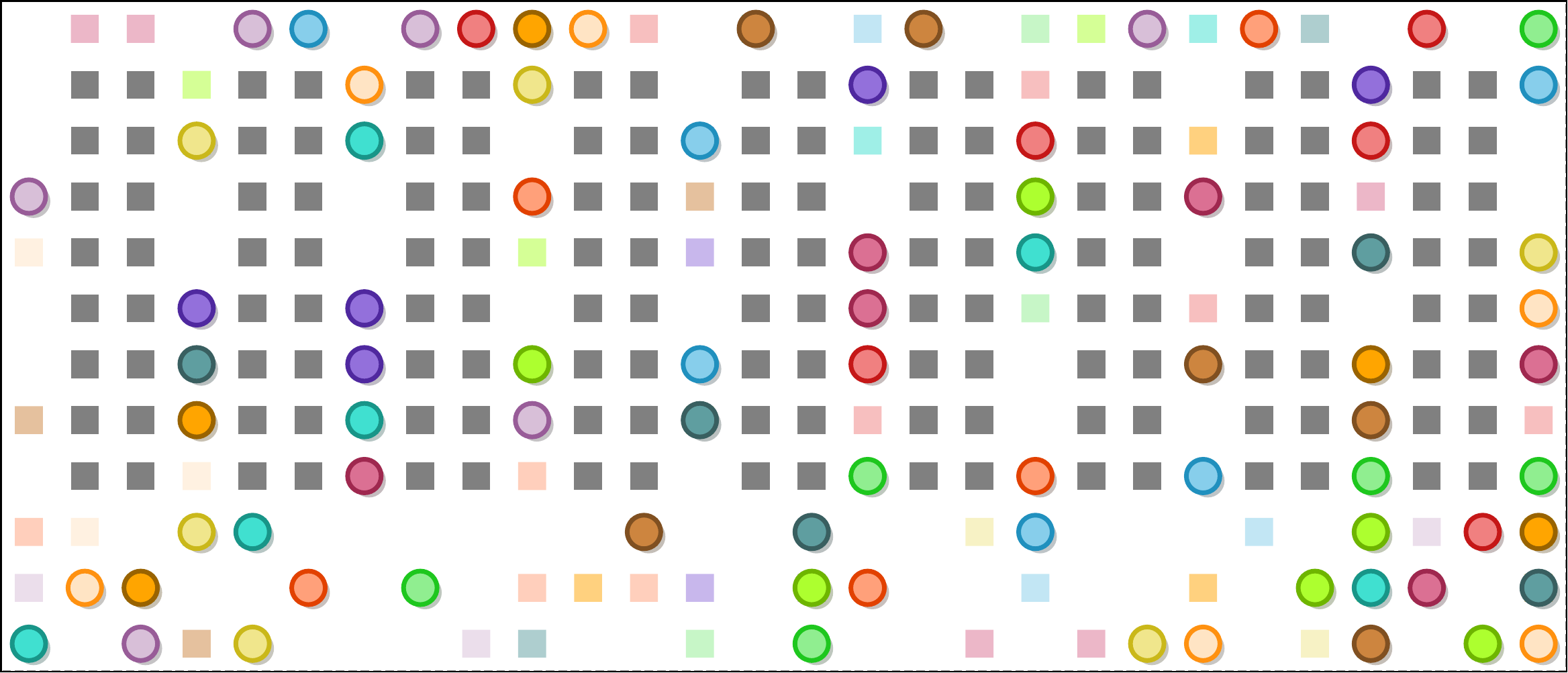}
         \caption{}
         \label{fig:eval-medium}
     \end{subfigure}
     \hfill
     \begin{subfigure}[b]{0.75\columnwidth}
         \centering
         \includegraphics[width=\columnwidth]{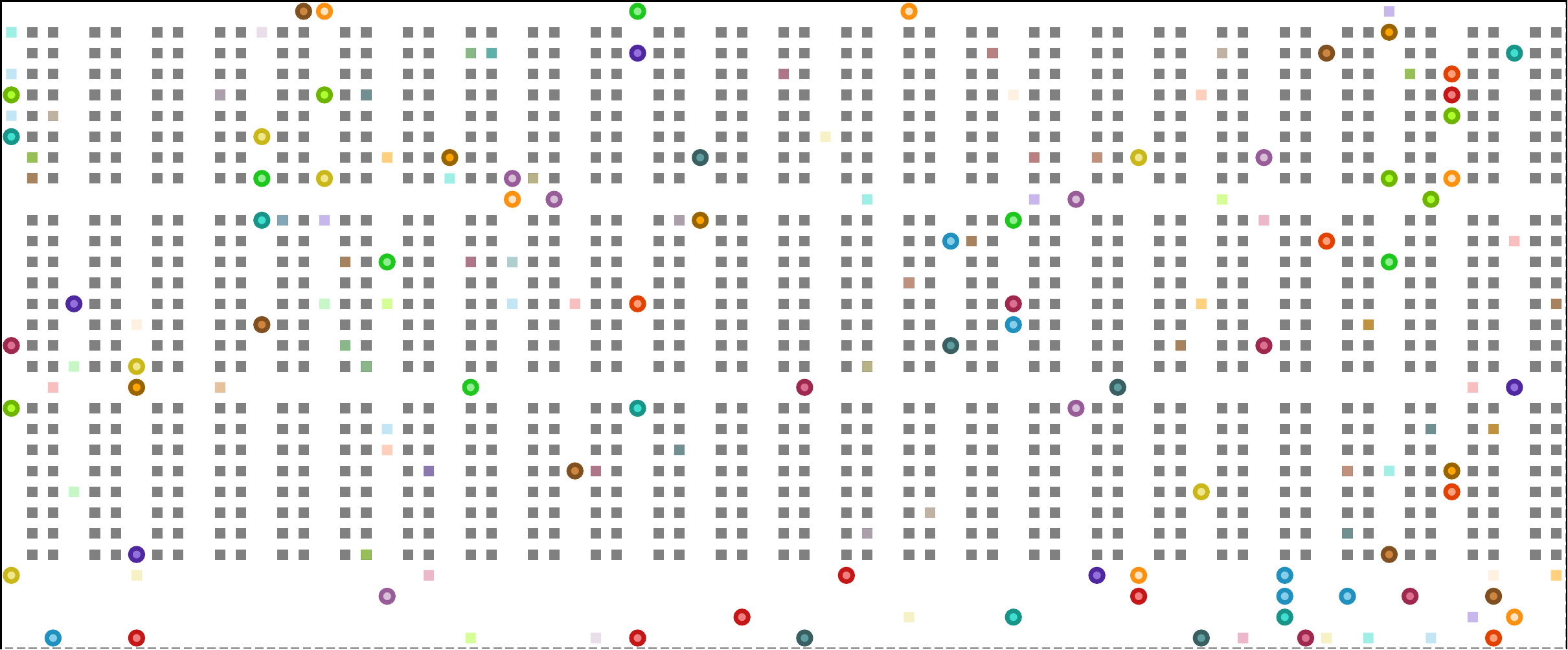}
         \caption{}
         \label{fig:eval-large}
     \end{subfigure}
     \hfill
     \begin{subfigure}[b]{0.75\columnwidth}
         \centering
         \includegraphics[width=\columnwidth]{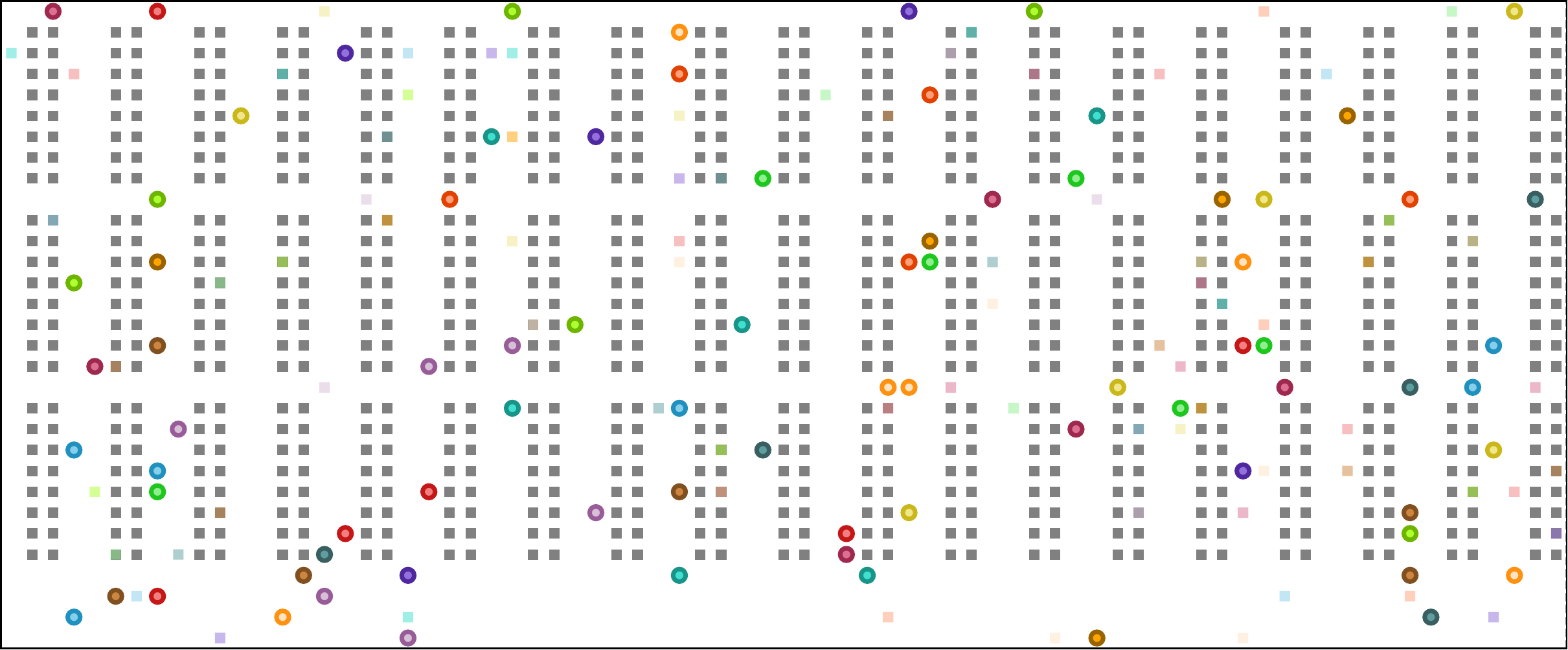}
         \caption{}
         \label{fig:eval-large2wide}
     \end{subfigure}
     \captionsetup{belowskip=-8pt,aboveskip=8pt}
    \caption{
        Warehouses used in our empirical evaluation.
        \protect{(\subref{fig:eval-medium})},
        \protect{(\subref{fig:eval-large})} and 
        \protect{(\subref{fig:eval-large2wide})} depict the \mediumEnv, \largeEnv and \largeEnvWide environments, 
        with~$80$ task agents (colorful dots).
        Here, rows of pods (gray rectangles) form long narrow aisles
        and goals (colorful rectangles) are selected from nearby workstations around the borders of each map.
    }
    % \vspace{-5mm}
    \label{fig:eval-warehouse}
\end{figure*}
\subsection{Additional Details}
\label{subsec:alg-details}
We give additional details on algorithmic building blocks described in Sec.~\ref{subsec:approach} and an example of the algorithm's flow.

\paragraph{Affected agents.}
The function \rm{getAffectedAgents} (invoked in Alg.~\ref{alg:tf-mapd}, Line~$7$% 
\arxiv{and detailed in Alg.~\ref{alg:get-affected})}{} 
computes the set of agents~$\tilde{\A}$ that are affected by the disruptions.
An agent is considered to be affected by a disruption if 
(i)~their path is blocked by a disruption
or if
(ii)~an agent with higher priority is affected by the disruption.
The second condition is important to allow agents whose path was constrained by higher-priority agents to re-plan their path after the higher-priority agents replanned their own paths.

\arxiv{
This set can be computed straightforwardly as outlined in Alg.~\ref{alg:get-affected} that receives as input the set of previously-computed paths~$\boldsymbol{\pi}$ and the set of active disruption locations~$\V_{\D}$.
The algorithm starts (Line~$1$) by adding all agents $a_i \in \A$ whose path $\pi_i\in\boldsymbol{\pi}$ intersects a vertex in~$\V_{\D}$.
All these agents are pushed into a stack $S$ (Line~$2$) and as long as $S$ is not empty (Lines~$3$-$8$), the algorithm pops an agent $a$, collects all agents with higher priority and if they are not in $\tilde{\A}$, adds them to $\tilde{\A}$ and to the stack $S$.
}
{
This set can be computed straightforwardly by adding all agents $a_i \in \A$ whose path $\pi_i\in\boldsymbol{\pi}$ intersects a vertex in~$\V_{\D}$.
All these agents are pushed into a stack $S$ and as long as $S$ is not empty, the algorithm pops an agent $a$, collects all agents with higher priority and if they are not in $\tilde{\A}$, adds them to $\tilde{\A}$ and to the stack $S$.
}

\arxiv{
\paragraph{Optimization---Resolving trapped agents.}
We define a \emph{trapped agent} as an agent that cannot reach its goal vertex due to being blocked by disruptions. 
As \wpbs only computes paths up to a limited planning window $\omega$, a trapped agent will be typically instructed to wait for the entire span of $\omega$ timesteps and will often not seek an alternative route. Moreover, this may happen over and over again for multiple planning iterations.
Following this observation we add the following optimization:
after disruptions are observed, we detect all trapped agents (this can be done by, e.g., running a Depth First Search (\algname{DFS}) on the graph from every agent's location treating disruptions as permanent obstacles).
If a trapped agent was detected, we commit to the plan that was found using terraforming (Line~$10$, Alg.~\ref{alg:tf-mapd}), instead of taking the path with the lower flowtime (Line~$11$). Specifically, in such cases of trapped agents, Line~$11$ is replaced with an assignment of $\boldsymbol{\pi_{\textsc{terra}}} \gets \boldsymbol{\pi}$. 
}{}

\paragraph{Example.}
We provide in Fig.~\ref{fig:MAPD_example} a visualization of \trhcr.
\textbf{Fig.~\ref{fig:tMAPD1}}: at time $t$ we have two task agents (\textcolor{teal}{$a_2$} and \textcolor{magenta}{$a_4$}) and  two free agents (\textcolor{blue}{$a_1$} and \textcolor{red}{$a_3$}).
Agents \textcolor{blue}{$a_1$} and \textcolor{red}{$a_3$} are assigned new tasks (Alg.~\ref{alg:tf-mapd}, Line~3) and \wpbs is used to plan suitable paths (Alg.~\ref{alg:tf-mapd}, Line~4). 
For each agent, we trace its path to its next goal (pick up $p_i$ or delivery $d_i$). 
\textbf{Fig.~\ref{fig:tMAPD2}}:  at time $t+2$ (i.e., after two timesteps), a disruption is detected (Alg.~\ref{alg:tf-mapd}, Line~5) due to an item dropped by agent~\textcolor{magenta}{$a_4$} along its path.
The affected agents (Line~7) are \textcolor{blue}{$a_1$} and \textcolor{teal}{$a_2$}, as the disruption blocks their paths.
We then use \wpbs to replan paths for the affected agents (Alg.~\ref{alg:tf-mapd}, Line~8) which would produce lengthy detours around the disruption.
Subsequently, a set of candidate obstacles to be moved (Line~9) is computed for ~$r=3$, illustrated with bold boundary. Then follows a call to getTerraformingTasks (Line~10). 
\textbf{Fig.~\ref{fig:tMAPD3}}: to compute the terraforming tasks (Alg.~\ref{alg:terra-tasks}), we form a \tmapf instance that includes the affected agents and the (self-propelled) candidate obstacles (Line~1). Paths computed are traced with a line. Note that  one self-propelled obstacle $o_1$ moved which is  returned by getObstaclesToMove (Line~2). Next, we create a terraforming task (Line~3), where the pickup location is the location of obstacle $o_1$ and the delivery location is the closest location that is reserved for terraforming: the reserved locations~$\hat{\V}$, are highlighted in orange at the top  of the map.
\textbf{Fig.~\ref{fig:tMAPD4}}:
Returning to Alg.~\ref{alg:tf-mapd}, Line~10 with the new terraforming task (Alg.~\ref{alg:terra-tasks}), we assign this new task to agent \textcolor{red}{$a_3$} (Line~11) hence postponing its current task. The new plan ~$\boldsymbol{\pi_{\textsc{terra}}}$ (Line~12) is outlined with the extraction of the obstacle and agents making use of the opening.
\textbf{Fig.~\ref{fig:tMAPD5}}:
At time $t+5$ (after three additional timesteps), agents \textcolor{blue}{$a_1$}, \textcolor{teal}{$a_2$} and \textcolor{red}{$a_3$} continue towards their delivery location. Agent \textcolor{red}{$a_3$} carries the obstacle away to maintain existing pathways clear and keep the shortcut open. Agent \textcolor{magenta}{$a_4$} reaches its respective task-delivery location and its next goal will be returning the obstacle's pickup location.

\section{Evaluation}
\label{sec:eval}
We evaluate our approach using maps inspired by autonomous warehouses: 
a \mediumEnv $24 \times 47$ map~\cite{felner2018adding,li2019disjoint}, a \largeEnv $32\times 75$ map~\cite{li2020new} and a \textsc{Large2Wide}\xspace $32\times 75$ map similar to \largeEnv but with corridors that are $2$-obstacle wide (see Fig.~\ref{fig:eval-warehouse}).

Unless stated otherwise, for each map we
(i)~run 25 random instances each with $600$ randomly generated pickup and delivery tasks,
(ii)~use terraforming radius of $r=8$ and 100 agents,
(iii)~consider two causes of disruptions (an immobilized agent and a dropped item)  where each type may occur along an agent path with probability of $0.5\%$  per simulation step and block a location for a random time interval in the range of $[40,60]$ simulation steps.\footnote{These numbers were verified by Amazon representative in private communication as being realistic values.} 
Finally, our algorithms are implemented in Python and tested on an Ubuntu machine with 16GB RAM and a 2.7GHz Intel~i7 CPU.

\begin{figure*}[t]
    \centering
    \begin{subfigure}[t]{0.32\textwidth}
        \includegraphics[width=0.9\columnwidth]{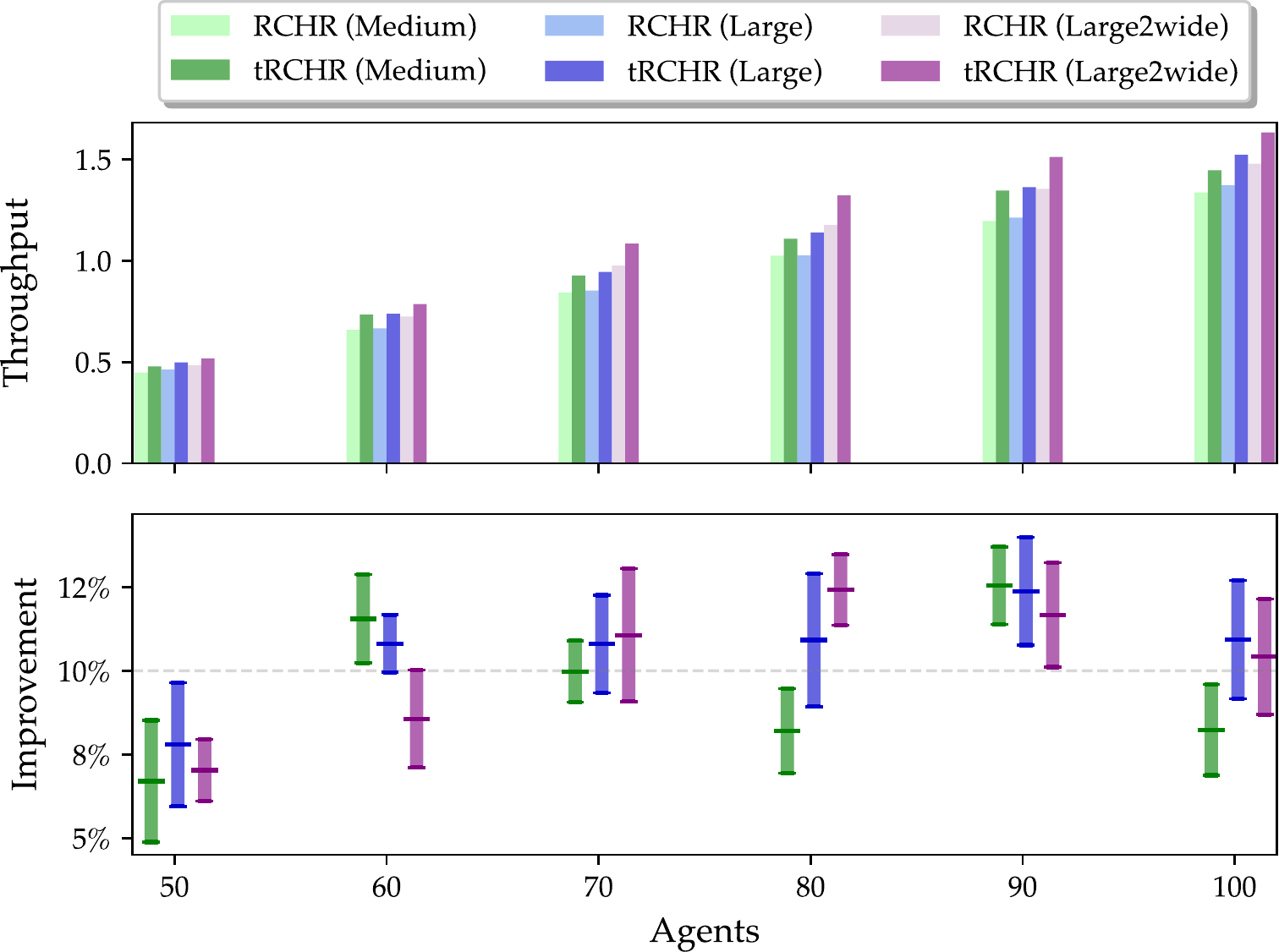}
         \caption{}
        \label{fig:eval-flowtime}
    \end{subfigure}
    \hfill
    \begin{subfigure}[t]{0.32\textwidth}
        \includegraphics[width=0.9\columnwidth]{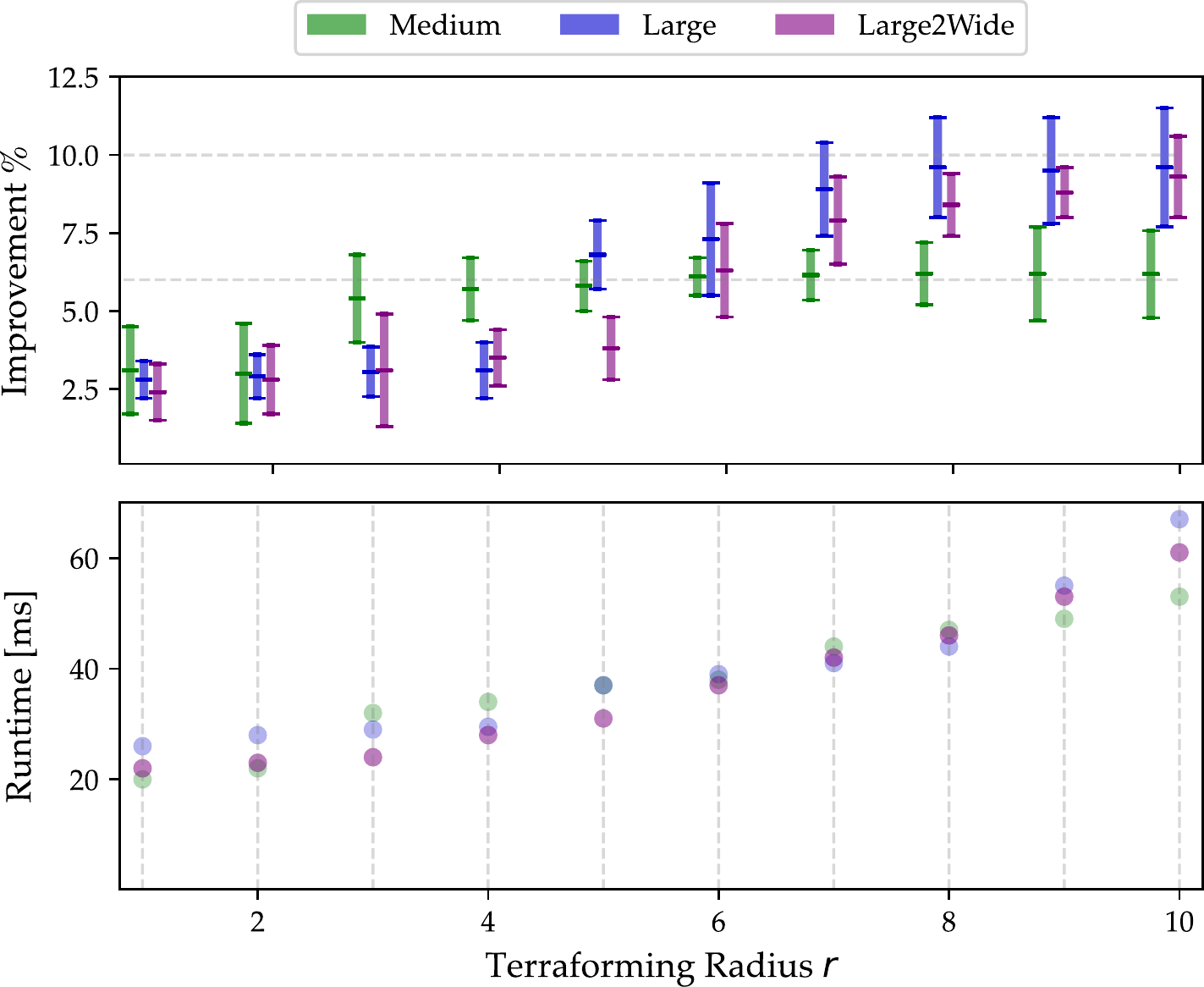}
        \caption{}
        \label{fig:eval-radius}
    \end{subfigure}
    \hfill
    \begin{subfigure}[t]{0.32\textwidth}
        \includegraphics[width=0.9\columnwidth]{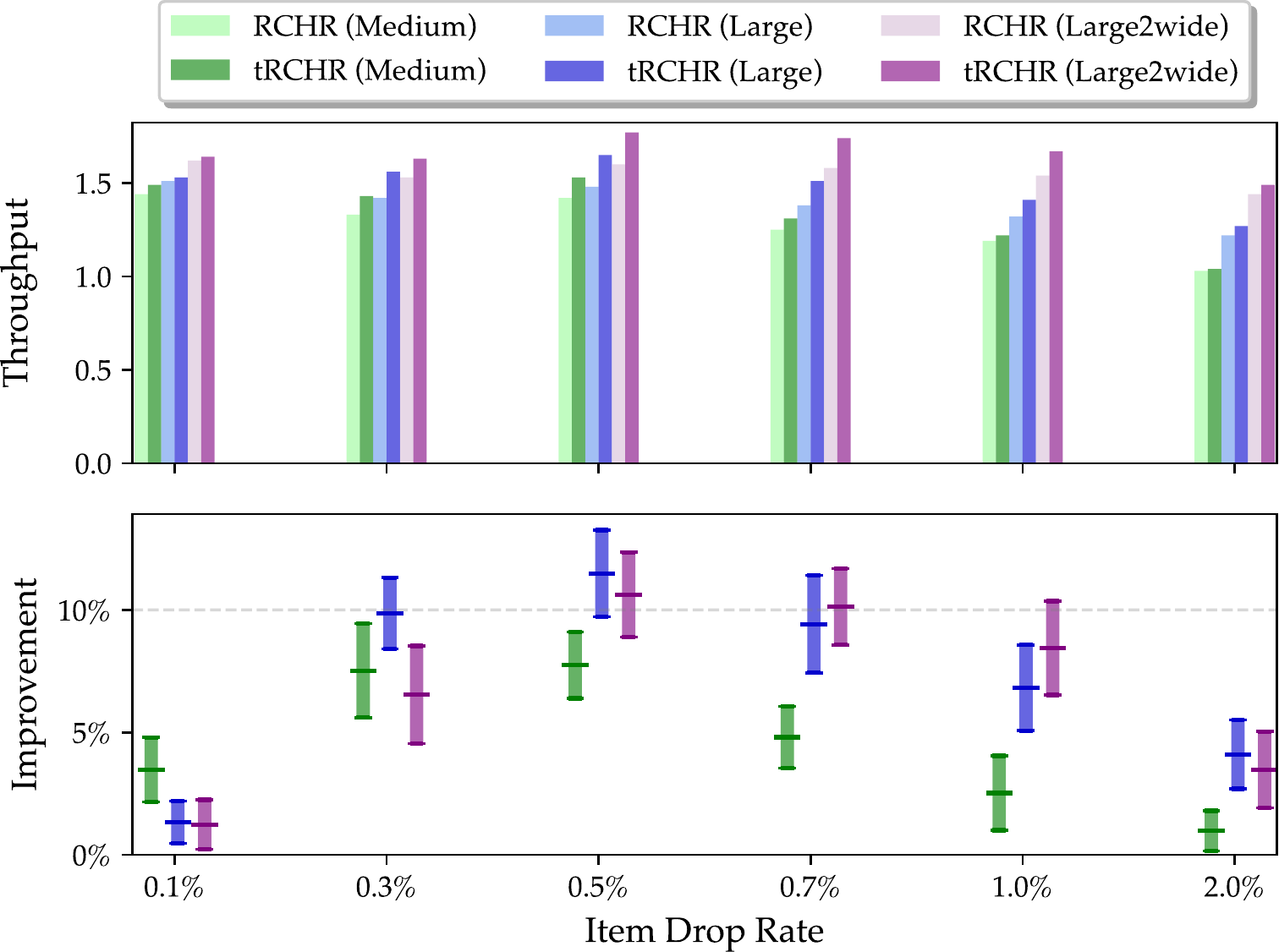}
        \caption{}
        \label{fig:eval-droprate}
    \end{subfigure}
    \captionsetup{belowskip=-8pt,aboveskip=3pt}
    \caption{
    (a)~Throughput of \mapd scenarios with disruptions, comparing the \rhcr-based baseline approach to \trhcr. 
    Top and bottom plots show the absolute throughput and improvement (i.e., ratio of \trhcr's throughput with the \rhcr-based throughput), respectively. Here, error bars denote one standard deviation.
    (b)~Evaluation of Terraforming radius $r$ on throughput improvement (top) and average runtime (bottom) per simulation timestep of \trhcr. 
    Shaded margins in the top figure corresponding to one standard deviation.
    (c)~Throughput as a function of disruption rate of dropped items.
    }
    % \vspace{-10pt}
    \label{fig:eval-results}
%    \vspace{-5mm}
\end{figure*}

\paragraph{Throughput.} 
We report a comparison of the throughput, which is the average number of tasks completed per simulation step, for \rhcr and \trhcr. 
We present (Fig.~\ref{fig:eval-flowtime}) the average throughput on each map for a varying  number of agents.
%$|\A| \in \{50, 60, 70, 80, 90, 100\}$.
%
Note that for all maps and values of $|\A|$, we obtain with terraforming an increased average throughput ranging between~$7-11\%$ and~$11\%$. 

\begin{figure}[h!]
     \centering
    \includegraphics[
    % trim={33mm 14mm 33mm 14mm}, clip,
    width=0.88\columnwidth]{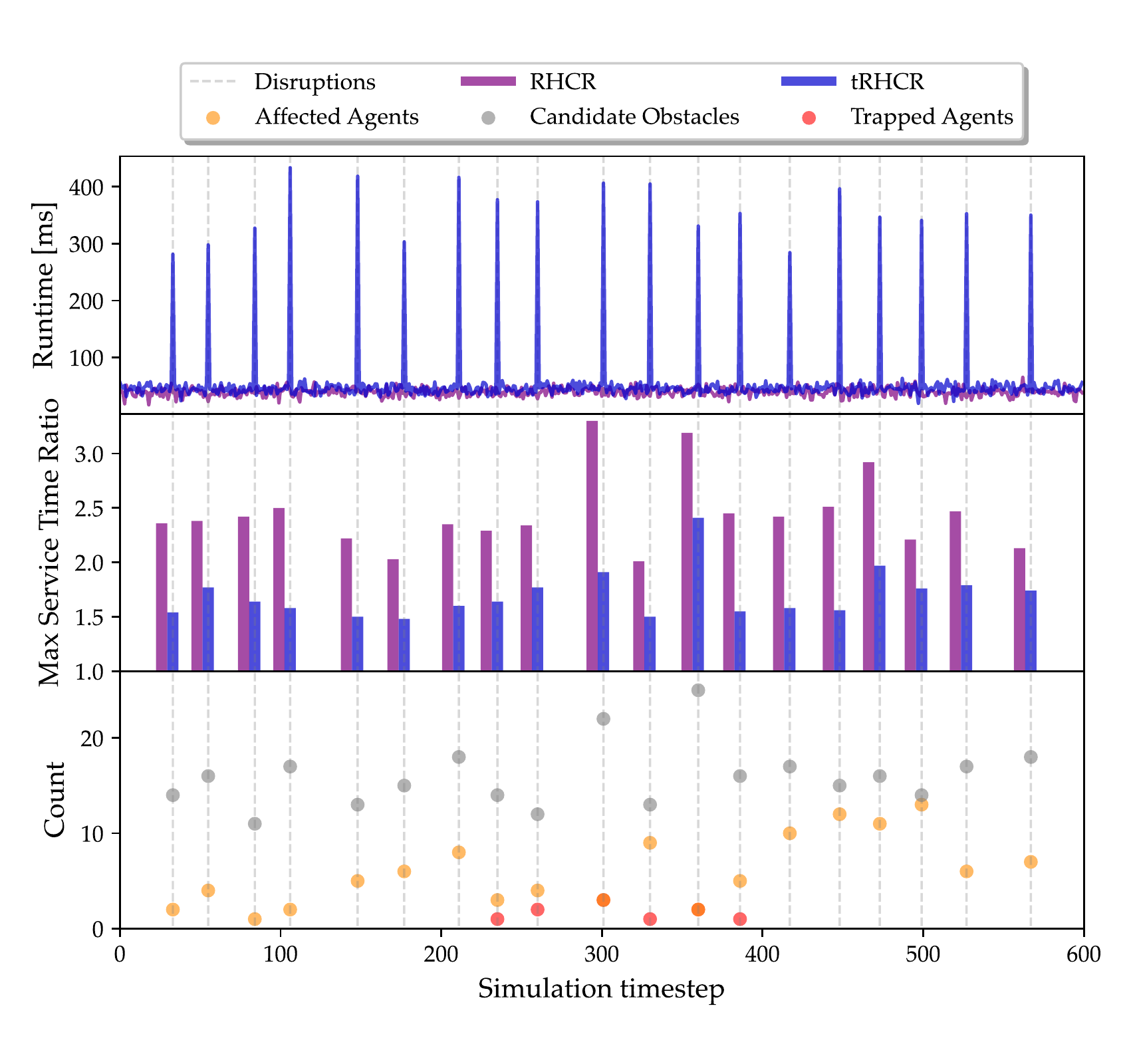}
        % \vspace{-2.5mm}
        \captionsetup{belowskip=-8pt,aboveskip=3pt}
    \caption{
        (Top)~Runtime (in ms) of \rhcr and \trhcr per simulation timestep with disruptions marked by dashed lines.
        \arxiv{
        (Middle)~The maximal task service time ratio for both \rhcr{} and \trhcr{} when disruptions occur (lower is better). Deep and light blue mark the performance of \trhcr with and without optimizing for trapped agents, as detailed in Sec.~\ref{subsec:alg-details}.}{
        (Middle)~The maximal task service time ratio for both \rhcr{} and \trhcr{} when disruptions occur (lower is better).
        }
        (Bottom)~The number of 
        agents affected by disruptions (Alg.~\ref{alg:tf-mapd}, Line~8), 
        the number of candidate obstacles submitted to \tfpbs{} (Alg.~\ref{alg:tf-mapd}, Line~9)
        and the number of trapped agents due to disruptions. 
}
\label{fig:eval-runtime}
% \vspace{-5mm}
\end{figure}

\paragraph{Runtime with and without disruptions.}
The lion's share of our \rhcr algorithms (with and without terraforming) is the \mapf planner which in our case is either \wpbs{} or \tfpbs{}.
To this end, we report (Fig.~\ref{fig:eval-runtime}, top) 
 the running time of each iteration of \rhcr and \trhcr, for a sample scenario with $100$ agents in the \largeEnv environment.

When no disruptions occur, the two algorithms are essentially identical and indeed their running time is roughly the same (it is not identical as terraforming causes agents' path to differ affecting future timesteps of the simulation). 
When disruptions occur (vertical dashed lines in Fig.~\ref{fig:eval-runtime}, top), then planning times for \tfpbs{} increases by an order of magnitude.
This is not surprising as the number of agents (including candidate obstacles as ``demi-agents'') that \tfpbs{} considers (which is the number of affected agents and candidate obstacles) is substantially larger than the number of agents that \wpbs{} considers (which is only the number of affected agents).
This can be seen in the bottom of Fig.~\ref{fig:eval-runtime} in which \tfpbs{} is applied on up to $10\times$ more agents than \wpbs.
%Indeed, a recent analysis \cite{gordon2021revisiting} of \mapf algorithms demonstrated that in certain settings, the algorithm's running time may be exponential in the number of agents. \KS{AFAIK, this analysis is for CBS, rather than a general statement about MAPF complexity, which is what we want. It might use a more subtle description.}
%
Having said that, when amortizing the running time of \tfpbs{} over the entire simulation, the average computation time, is still within the arguably tolerable range of $40ms$ per planning window, compared with the average runtime of \wpbs of~$5ms$.
Moreover, the improvement in throughput as previously discussed and demonstrated in Fig.~\ref{fig:eval-flowtime} can make the extra runtime worthwhile.

% Note: Avoid trim/clip for camera ready

\paragraph{Maximum task service time ratio.}
Recall that the service time is the number of timesteps elapsed between a task's pick-up time and its completion (drop-off) time, whereas the ideal service time is the shortest service time attainable for a task when no other agents (or disruptions) interfere with its execution. For each task, the ratio between the former and the latter is a measure of task delay. 
We compare (Fig.~\ref{fig:eval-runtime}, middle) the maximal service time ratio for both \rhcr{} and \trhcr{}.
Note that Fig.~\ref{fig:eval-runtime} shows the service time of the most-delayed task, while the average tasks' delay is reflected in the throughout as shown in Fig.~\ref{fig:eval-flowtime}.
This allows to pinpoint where the increase in throughput (Fig.~\ref{fig:eval-flowtime}) comes from and to justify the need to perform the expensive operation of terraforming when disruptions occur (Fig.~\ref{fig:eval-runtime}, top). 

Indeed, we can see that the maximal service time ratio for \rhcr{} is roughly $2\times$ to $3\times$ larger than the maximal service time ratio of \trhcr{}.
This is expected as agents that are blocked by disruptions require detouring obstructed vertices and in doing so increase their service time. To a greater extent, agents that become trapped by surrounding disruptions can only wait in place until an opening is cleared. In contrast, terraforming creates pathways (when beneficial) and shortcuts that decrease the service time of blocked agents.

\arxiv{
Regarding the  optimization where we account for trapped agents,
when there are no trapped agents, both variants perform exactly the same (see behaviour for the first $200$ simulation timesteps in Fig.~\ref{fig:eval-runtime}).
However, when trapped agents exist, the maximal task service time increases dramatically unless an evacuation route is intentionally cleared through terraforming.  
Note that after the first instance where trapped agents exist, the rest of the simulation differs which is why in the last timesteps, even though there are no trapped agents, the maximal task service time of the two variants differs.
}{}
\paragraph{Terraforming radius evaluation.} Recall that the terraforming radius $r$  is used to compute the set of candidate movable obstacles~$\tilde{\O}$ (Alg.~\ref{alg:tf-mapd}, Line~9).
The number of candidate movable obstacles~$\vert\tilde{\O}\vert$, and thus the complexity of \tfpbs, grows proportionally with~$r$.
However, the potential improvement in solution quality grows with $r$ as well since terraforming has more options to displace obstacles.
Thus, we evaluate both the throughput improvement (Fig.~\ref{fig:eval-radius}, top) and the algorithm's runtime (Fig.~\ref{fig:eval-radius}, bottom) as a function of the terraforming radius.
We can  see the tradeoff between the increased computation time and the improved throughput that happens as $r$ increases. Empirically $r=8$  balances between the two in all environments.

\paragraph{Evaluating item drop rate.}
We evaluate the effect of disruption rates on  throughput.
Specifically, we consider varying rates of disruptions caused by a dropped item, and maintain agent breakdown rate of $0.5\%$ per timestep
(we fix disruption rates caused by agent breakdowns because higher disruption rates do not leave enough agents to perform terraforming and to complete tasks).
Results, summarized in Fig.~\ref{fig:eval-droprate} show that as disruption rates caused by dropped items increase, so does the throughput improvement, which peaks at $0.5\%$ disruptions per timestep. This improvement then decreases as
(i)~there may be more disruptions than can be handled by free agents
or
(ii)~alternative paths offered by obstacle displacements become blocked as well.
%

% We evaluate $100$ agents ($25$ seeds) for varying rates of disruptions caused by a dropped item, and maintain agent breakdown rate of $0.5\%$ per timestep.
% Figure \ref{fig:eval-droprate} shows an increasing benefit to Terraforming as the drop-rate approaches $0.5\%$.
% The medium warehouse exhibits more relative improvement, since disruptions cause blockages and a smaller map provides fewer alternative routes.
% The greatest benefit is achieved at $0.5\%$ disruption rate per timestep. Here, non-terraforming suffers while terraforming is able to maintain performance - as evident by the rate of improvement (bottom chart).
% Further increasing the item drop-rate reduces the benefit of Terraforming.
% This is because 1) there may be more disruptions than can be handled by free agents, or 2) obstacle evacuation paths are themselves blocked by disruptions, as well as surrounding agents.
% At the catastrophic ratio of $2.0\%$ item drop rate, we see that throughput degrades by $15\!-\!40\%$ compared with $0.1\%$ item drop rate.
% While Terraforming still offers a benefit, its efficacy is clearly constrained by the intense rate of disruptions.

% Note: Avoid trim/clip for camera ready

\section{Discussion and Future Work}
\label{sec:discussion}
% \modified{The idea of intelligent agents terraforming their environment has intrigued humanity for nearly a century. Although originally pertaining to space exploration, terraforming may very soon become a reality here on earth. For example, contemporary research into construction sites of the future examines autonomous excavators \cite{liu2022object}, trucks and dozers, working together as a multi-agent system that reshapes and forms the surrounding environment towards a common goal.}

In this work we explored the potential benefits of terraforming within \mapd, with emphasis on mitigating the impact of unforeseen disruptions\arxiv{ that obstruct critical pathways and force agents to use lengthy detours.}{.} The extreme case of an agent becoming trapped by disruptions\arxiv{in its vicinity}{} offers the most convincing motivation for terraforming. Without terraforming, such agents cannot make any progress while  terraforming allows to maintain (and even improve) throughput by having nearby agents create shortcuts and reduce per-task execution time.

To harness the full potential of \tmapf, we envision its application to \mapd where agents en-route to collect an item can optimize the environment through local changes that serve multiple agents, regardless of disruptions. As future work we wish to consider an approach that facilitates subtle environment manipulation with little associated cost, but with enough foresight as to yield a significant benefit to nearby agents, which could provide a substantial boost to overall throughput. 
Furthermore, we plan to formally analyze the new variant's complexity in future work.

\arxiv{Finally, we foresee applications of our work beyond our motivating example of \mapd in warehouses.
For example, multi-agent systems conducting search and rescue missions will likely need to coordinate clearing debris, opening pathways and rescuing subjects that are initially inaccessible. Some tasks may require multiple agents to complete, perhaps agents with different expertise for various roles, with their success dependent on their ability to leverage the environment to the best of their capacity.}{}

\bibliography{aaai23.bib}

\begin{thebibliography}{27}
\providecommand{\natexlab}[1]{#1}

\bibitem[{Atzmon et~al.(2018)Atzmon, Stern, Felner, Wagner, Bart{\'a}k, and
  Zhou}]{atzmon2018robust}
Atzmon, D.; Stern, R.; Felner, A.; Wagner, G.; Bart{\'a}k, R.; and Zhou, N.-F.
  2018.
\newblock Robust multi-agent path finding.
\newblock In \emph{{Symposium on Combinatorial Search ({SoCS})}}.

\bibitem[{Atzmon et~al.(2020)Atzmon, Stern, Felner, Wagner, Bart{\'a}k, and
  Zhou}]{atzmon2020robust}
Atzmon, D.; Stern, R.; Felner, A.; Wagner, G.; Bart{\'a}k, R.; and Zhou, N.-F.
  2020.
\newblock Robust multi-agent path finding and executing.
\newblock \emph{Journal of Artificial Intelligence Research}, 67: 549--579.

\bibitem[{Barer et~al.(2014)Barer, Sharon, Stern, and Felner}]{BSSF14}
Barer, M.; Sharon, G.; Stern, R.; and Felner, A. 2014.
\newblock Suboptimal Variants of the Conflict-Based Search Algorithm for the
  Multi-Agent Pathfinding Problem.
\newblock In \emph{{European Conference on Artificial Intelligence ({ECAI})}},
  volume 263, 961--962.

\bibitem[{Bellusci, Basilico, and Amigoni(2020)}]{bellusci2020multi}
Bellusci, M.; Basilico, N.; and Amigoni, F. 2020.
\newblock Multi-Agent Path Finding in Configurable Environments.
\newblock In \emph{{Autonomous Agents and MultiAgent Systems ({AAMAS})}},
  159--167.

\bibitem[{Belov et~al.(2020)Belov, Du, de~la Banda, Harabor, Koenig, and
  Wei}]{BelovDBHKW20}
Belov, G.; Du, W.; de~la Banda, M.~G.; Harabor, D.; Koenig, S.; and Wei, X.
  2020.
\newblock From Multi-Agent Pathfinding to 3D Pipe Routing.
\newblock In Harabor, D.; and Vallati, M., eds., \emph{{Symposium on
  Combinatorial Search ({SoCS})}}, 11--19.

\bibitem[{Boyarski et~al.(2015)Boyarski, Felner, Stern, Sharon, Tolpin,
  Betzalel, and Shimony}]{BFSSTBS15}
Boyarski, E.; Felner, A.; Stern, R.; Sharon, G.; Tolpin, D.; Betzalel, O.; and
  Shimony, S.~E. 2015.
\newblock {ICBS:} Improved Conflict-Based Search Algorithm for Multi-Agent
  Pathfinding.
\newblock In \emph{{International Joint Conferences on Artificial Intelligence
  ({IJCAI})}}, 740--746.

\bibitem[{Choudhury et~al.(2021)Choudhury, Solovey, Kochenderfer, and
  Pavone}]{Choudhury.ea.21}
Choudhury, S.; Solovey, K.; Kochenderfer, M.~J.; and Pavone, M. 2021.
\newblock Efficient Large-Scale Multi-Drone Delivery using Transit Networks.
\newblock \emph{{Journal of Artificial Intelligence Research}}, 70: 757--788.

\bibitem[{Erdem et~al.(2013)Erdem, Kisa, Oztok, and Sch{\"u}ller}]{EKOS13}
Erdem, E.; Kisa, D.~G.; Oztok, U.; and Sch{\"u}ller, P. 2013.
\newblock A general formal framework for pathfinding problems with multiple
  agents.
\newblock In \emph{{Association for the Advancement of Artificial Intelligence
  ({AAAI})}}.

\bibitem[{Felner et~al.(2018)Felner, Li, Boyarski, Ma, Cohen, Kumar, and
  Koenig}]{felner2018adding}
Felner, A.; Li, J.; Boyarski, E.; Ma, H.; Cohen, L.; Kumar, T.~S.; and Koenig,
  S. 2018.
\newblock Adding heuristics to conflict-based search for multi-agent path
  finding.
\newblock \emph{{ International Conference on Automated Planning and Scheduling
  (ICAPS)}}, 28: 83--87.

\bibitem[{Greshler et~al.(2021)Greshler, Gordon, Salzman, and Shimkin}]{GGSS21}
Greshler, N.; Gordon, O.; Salzman, O.; and Shimkin, N. 2021.
\newblock Cooperative Multi-Agent Path Finding: Beyond Path Planning and
  Avoidance.
\newblock In \emph{{Symposium on Multi-Robot and Multi-Agent Systems ({MRS})}},
  20--28.

\bibitem[{Li et~al.(2021{\natexlab{a}})Li, Chen, Zheng, Chan, Harabor, Stuckey,
  Ma, and Koenig}]{LCZCHS0K21}
Li, J.; Chen, Z.; Zheng, Y.; Chan, S.; Harabor, D.; Stuckey, P.~J.; Ma, H.; and
  Koenig, S. 2021{\natexlab{a}}.
\newblock Scalable Rail Planning and Replanning: Winning the 2020 Flatland
  Challenge.
\newblock In \emph{{ International Conference on Automated Planning and
  Scheduling (ICAPS)}}, 477--485.

\bibitem[{Li et~al.(2020)Li, Gange, Harabor, Stuckey, Ma, and
  Koenig}]{li2020new}
Li, J.; Gange, G.; Harabor, D.; Stuckey, P.~J.; Ma, H.; and Koenig, S. 2020.
\newblock New techniques for pairwise symmetry breaking in multi-agent path
  finding.
\newblock In \emph{{ International Conference on Automated Planning and
  Scheduling (ICAPS)}}, volume~30, 193--201.

\bibitem[{Li et~al.(2019)Li, Harabor, Stuckey, Felner, Ma, and
  Koenig}]{li2019disjoint}
Li, J.; Harabor, D.; Stuckey, P.~J.; Felner, A.; Ma, H.; and Koenig, S. 2019.
\newblock Disjoint splitting for multi-agent path finding with conflict-based
  search.
\newblock In \emph{{ International Conference on Automated Planning and
  Scheduling (ICAPS)}}, volume~29, 279--283.

\bibitem[{Li et~al.(2021{\natexlab{b}})Li, Tinka, Kiesel, Durham, Kumar, and
  Koenig}]{li2021lifelong}
Li, J.; Tinka, A.; Kiesel, S.; Durham, J.; Kumar, S.; and Koenig, S.
  2021{\natexlab{b}}.
\newblock Lifelong Multi-Agent Path Finding in Large-Scale Warehouses.
\newblock In \emph{{Association for the Advancement of Artificial Intelligence
  ({AAAI})}}.

\bibitem[{Liu et~al.(2019)Liu, Ma, Li, and Koenig}]{liu2019task}
Liu, M.; Ma, H.; Li, J.; and Koenig, S. 2019.
\newblock Task and path planning for multi-agent pickup and delivery.
\newblock \emph{{Autonomous Agents and MultiAgent Systems ({AAMAS})}}, 12:
  206--208.

\bibitem[{Ma et~al.(2019{\natexlab{a}})Ma, Harabor, Stuckey, Li, and
  Koenig}]{ma2019searching}
Ma, H.; Harabor, D.; Stuckey, P.~J.; Li, J.; and Koenig, S. 2019{\natexlab{a}}.
\newblock Searching with consistent prioritization for multi-agent path
  finding.
\newblock In \emph{{Association for the Advancement of Artificial Intelligence
  ({AAAI})}}, volume~33, 7643--7650.

\bibitem[{Ma et~al.(2019{\natexlab{b}})Ma, H{\"o}nig, Kumar, Ayanian, and
  Koenig}]{ma2019lifelong}
Ma, H.; H{\"o}nig, W.; Kumar, T.~S.; Ayanian, N.; and Koenig, S.
  2019{\natexlab{b}}.
\newblock Lifelong path planning with kinematic constraints for multi-agent
  pickup and delivery.
\newblock In \emph{{Association for the Advancement of Artificial Intelligence
  ({AAAI})}}, volume~33, 7651--7658.

\bibitem[{Ma et~al.(2017)Ma, Li, Kumar, and Koenig}]{ma2017lifelong}
Ma, H.; Li, J.; Kumar, T. K.~S.; and Koenig, S. 2017.
\newblock Lifelong multi-agent path finding for online pickup and delivery
  tasks.
\newblock In \emph{{Autonomous Agents and MultiAgent Systems ({AAMAS})}},
  837--845.

\bibitem[{Madar, Solovey, and Salzman(2022)}]{MadarSS22}
Madar, N.; Solovey, K.; and Salzman, O. 2022.
\newblock Leveraging Experience in Lifelong Multi-Agent Pathfinding.
\newblock In \emph{{Symposium on Combinatorial Search ({SoCS})}}, 118--126.

\bibitem[{Mohanty et~al.(2020)Mohanty, Nygren, Laurent, Schneider, Scheller,
  Bhattacharya, Watson, Egli, Eichenberger, Baumberger
  et~al.}]{mohanty2020flatland}
Mohanty, S.; Nygren, E.; Laurent, F.; Schneider, M.; Scheller, C.;
  Bhattacharya, N.; Watson, J.; Egli, A.; Eichenberger, C.; Baumberger, C.;
  et~al. 2020.
\newblock Flatland-rl: Multi-agent reinforcement learning on trains.
\newblock \emph{arXiv preprint arXiv:2012.05893}.

\bibitem[{Okumura, Tamura, and Défago(2021)}]{okumura2021iterative}
Okumura, K.; Tamura, Y.; and Défago, X. 2021.
\newblock Iterative Refinement for Real-Time Multi-Robot Path Planning.
\newblock In \emph{2021 IEEE/RSJ International Conference on Intelligent Robots
  and Systems (IROS)}, 9690--9697.

\bibitem[{Salzman and Stern(2020)}]{SalzmanS20}
Salzman, O.; and Stern, R. 2020.
\newblock Research Challenges and Opportunities in Multi-Agent Path Finding and
  Multi-Agent Pickup and Delivery Problems.
\newblock In \emph{{Autonomous Agents and MultiAgent Systems ({AAMAS})}},
  1711--1715.

\bibitem[{Sharon et~al.(2015)Sharon, Stern, Felner, and
  Sturtevant}]{sharon2015conflict}
Sharon, G.; Stern, R.; Felner, A.; and Sturtevant, N.~R. 2015.
\newblock Conflict-based search for optimal multi-agent pathfinding.
\newblock \emph{Artificial Intelligence}, 219: 40--66.

\bibitem[{Stern et~al.(2019)Stern, Sturtevant, Felner, Koenig, Ma, Walker, Li,
  Atzmon, Cohen, Kumar et~al.}]{stern2019multi}
Stern, R.; Sturtevant, N.; Felner, A.; Koenig, S.; Ma, H.; Walker, T.; Li, J.;
  Atzmon, D.; Cohen, L.; Kumar, S.; et~al. 2019.
\newblock Multi-Agent Pathfinding: Definitions, Variants, and Benchmarks.
\newblock \emph{{Symposium on Combinatorial Search ({SoCS})}}, 10: 151--158.

\bibitem[{Surynek et~al.(2016)Surynek, Felner, Stern, and Boyarski}]{SFSB16}
Surynek, P.; Felner, A.; Stern, R.; and Boyarski, E. 2016.
\newblock Efficient {SAT} approach to multi-agent path finding under the sum of
  costs objective.
\newblock In \emph{{European Conference on Artificial Intelligence ({ECAI})}},
  810--818. IOS Press.

\bibitem[{Wurman, D’Andrea, and Mountz(2008)}]{wurman2008cooperative}
Wurman, P.~R.; D’Andrea, R.; and Mountz, M. 2008.
\newblock Coordinating Hundreds of Cooperative, Autonomous Vehicles in
  Warehouses.
\newblock \emph{AI Magazine}, 29(1): 9.

\bibitem[{Yu and LaValle(2012)}]{YL12}
Yu, J.; and LaValle, S.~M. 2012.
\newblock Multi-agent Path Planning and Network Flow.
\newblock In \emph{{Workshop on the Algorithmic Foundations of Robotics
  (WAFR)}}, volume~86, 157--173. Springer.

\end{thebibliography}

\end{document}